\documentclass[acmsmall]{acmart}

\AtBeginDocument{%
  \providecommand\BibTeX{{%
    \normalfont B\kern-0.5em{\scshape i\kern-0.25em b}\kern-0.8em\TeX}}}

\usepackage{amsmath,amsfonts}
\usepackage{algorithmic}
\usepackage{algorithm}
\usepackage{array}
\usepackage{textcomp}
\usepackage{stfloats}
\usepackage{url}
\usepackage{tabularx}
\usepackage{verbatim}
\usepackage{graphicx}
\usepackage{nicefrac}
\usepackage{multirow}
\usepackage{subfigure}
\usepackage{float}
\usepackage{hyperref}
\usepackage{makecell}
\usepackage{graphicx}
\DeclareGraphicsExtensions{.pdf}
\usepackage{color}

\def\ie{i.e.}

\usepackage{newfloat}
\usepackage{listings}
\begin{document}

\title{Rank Aggregation in Crowdsourcing for Listwise Annotations}

\author{Wenshui Luo}
\authornote{The authors contributed equally to this research.}
\orcid{0009-0005-9283-740X}
\affiliation{%
  \institution{Nanjing University of Science and Technology}
  \city{Nanjing}
  \postcode{210094}
  \country{China}
}
\email{randylo@njust.edu.cn}

\author{Haoyu Liu}
\authornotemark[1]
\authornote{Correspondence authors.}
\affiliation{%
  \institution{Fuxi AI Lab, NetEase Games}
  \city{Hangzhou}
  \country{China}
  \postcode{310052}
}
\email{liuhaoyu03@corp.netease.com}

\author{Yongliang Ding}
\authornotemark[1]
\affiliation{%
  \institution{Nanjing University of Science and Technology}
  \city{Nanjing}
  \postcode{210094}
  \country{China}
}
\email{1242388760@njust.edu.cn}

\author{Tao Zhou}
\affiliation{%
  \institution{Nanjing University of Science and Technology}
  \city{Nanjing}
  \postcode{210094}
  \country{China}
}
\email{taozhou.ai@gmail.com}

\author{Sheng Wan}
\affiliation{%
  \institution{Nanjing University of Science and Technology}
  \city{Nanjing}
  \postcode{210094}
  \country{China}
}
\email{wansheng315@hotmail.com}

\author{Runze Wu}
\authornotemark[2]
\affiliation{%
  \institution{Fuxi AI Lab, NetEase Games}
  \city{Hangzhou}
  \country{China}
  \postcode{310052}
}
\email{wurunze1@corp.netease.com}

\author{Minmin Lin}
\affiliation{%
  \institution{Fuxi AI Lab, NetEase Games}
  \city{Hangzhou}
  \country{China}
  \postcode{310052}
}
\email{linminmin01@corp.netease.com}

\author{Cong Zhang}
\affiliation{%
  \institution{Fuxi AI Lab, NetEase Games}
  \city{Hangzhou}
  \country{China}
  \postcode{310052}
}
\email{zhangcong@corp.netease.com}

\author{Changjie Fan}
\affiliation{%
  \institution{Fuxi AI Lab, NetEase Games}
  \city{Hangzhou}
  \country{China}
  \postcode{310052}
}
\email{fanchangjie@corp.netease.com}

\author{Chen Gong}
\authornotemark[2]
\affiliation{%
  \institution{Nanjing University of Science and Technology}
  \city{Nanjing}
  \postcode{210094}
  \country{China}
}
\email{chen.gong@njust.edu.cn}

\renewcommand{\shortauthors}{Wenshui and Haoyu, et al.}

\begin{abstract}
  Rank aggregation through crowdsourcing has recently gained significant attention, particularly in the context of listwise ranking annotations. However, existing methods primarily focus on a single problem and partial ranks, while the aggregation of listwise full ranks across numerous problems remains largely unexplored. This scenario finds relevance in various applications, such as model quality assessment and reinforcement learning with human feedback. In light of practical needs, we propose LAC, a \underline{\textbf{L}}istwise rank \underline{\textbf{A}}ggregation method in \underline{\textbf{C}}rowdsourcing, where the global position information is carefully measured and included. In our design, an especially proposed annotation quality indicator is employed to measure the discrepancy between the annotated rank and the true rank. We also take the difficulty of the ranking problem itself into consideration, as it directly impacts the performance of annotators and consequently influences the final results. To our knowledge, LAC is the first work to directly deal with the full rank aggregation problem in listwise crowdsourcing, and simultaneously infer the difficulty of problems, the ability of annotators, and the ground-truth ranks in an unsupervised way. To evaluate our method, we collect a real-world business-oriented dataset for paragraph ranking. Experimental results on both synthetic and real-world benchmark datasets demonstrate the effectiveness of our proposed LAC method. 
\end{abstract}




\maketitle

\section{Introduction}

Recently, inferring ranking~\cite{group_event_recommendation, deep_partial_rank_aggregation, top_k_pairwise_crowdsourcing, better_algorithms_rank_aggregation, tkdd_crowdsourcing_truth_inference, tkdd_exploting_heterogeneous} over a set of items has gained increasing attention due to its wide range of applications, such as information retrieval~\cite{information_retrieval_system_1}, recommendation systems~\cite{recommendation_system_1}, and RLHF (Reinforcement Learning from Human Feedback) for finetuning large language models~\cite{RLHF_application}. This task aims to train a ranking model in a supervised way~\cite{learning_ranking_paper_1}, thereby requiring a large amount of well-annotated data. 

Unfortunately, it will be of high cost to hire expert annotators, especially when the scale of data is large. To accommodate this dilemma, many practitioners resort to crowdsourcing platforms~\cite{crowdsourcing_platform, RSWC_crowdsourcing, 7384520, ML_crowdsoucing}, such as Amazon Mechanical Turk and CrowdFlower. They distribute ranking problems into multiple sub-tasks, which are required to be solved by crowdsourced annotators. Aggregation methods are subsequently employed so that the solutions of sub-tasks can be aggregated and the true ranks of the original ranking problems can be derived. Among these procedures, the effectiveness of rank aggregation algorithms is of significant importance for the performance of the final aggregated results.

\begin{figure}[t]

  \centering
    \begin{minipage}{0.49\linewidth}
		\centering
		\subfigure[Previous rank aggregation settings.]
		{\includegraphics[width=\linewidth]{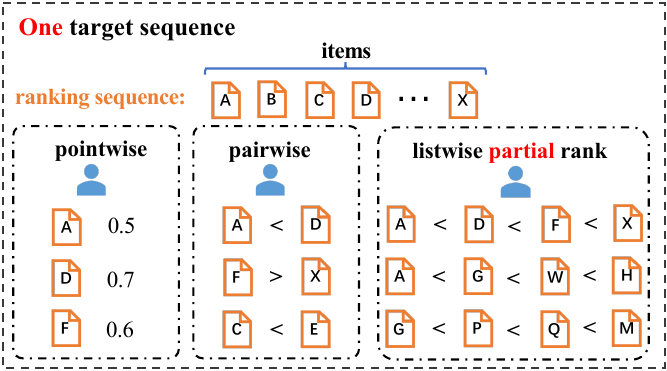}
  }
    \end{minipage}
    \hfill
    \begin{minipage}{0.49\linewidth}
		\centering
		\subfigure[The proposed listwise full rank aggregation task.]
		{\includegraphics[width=\linewidth]{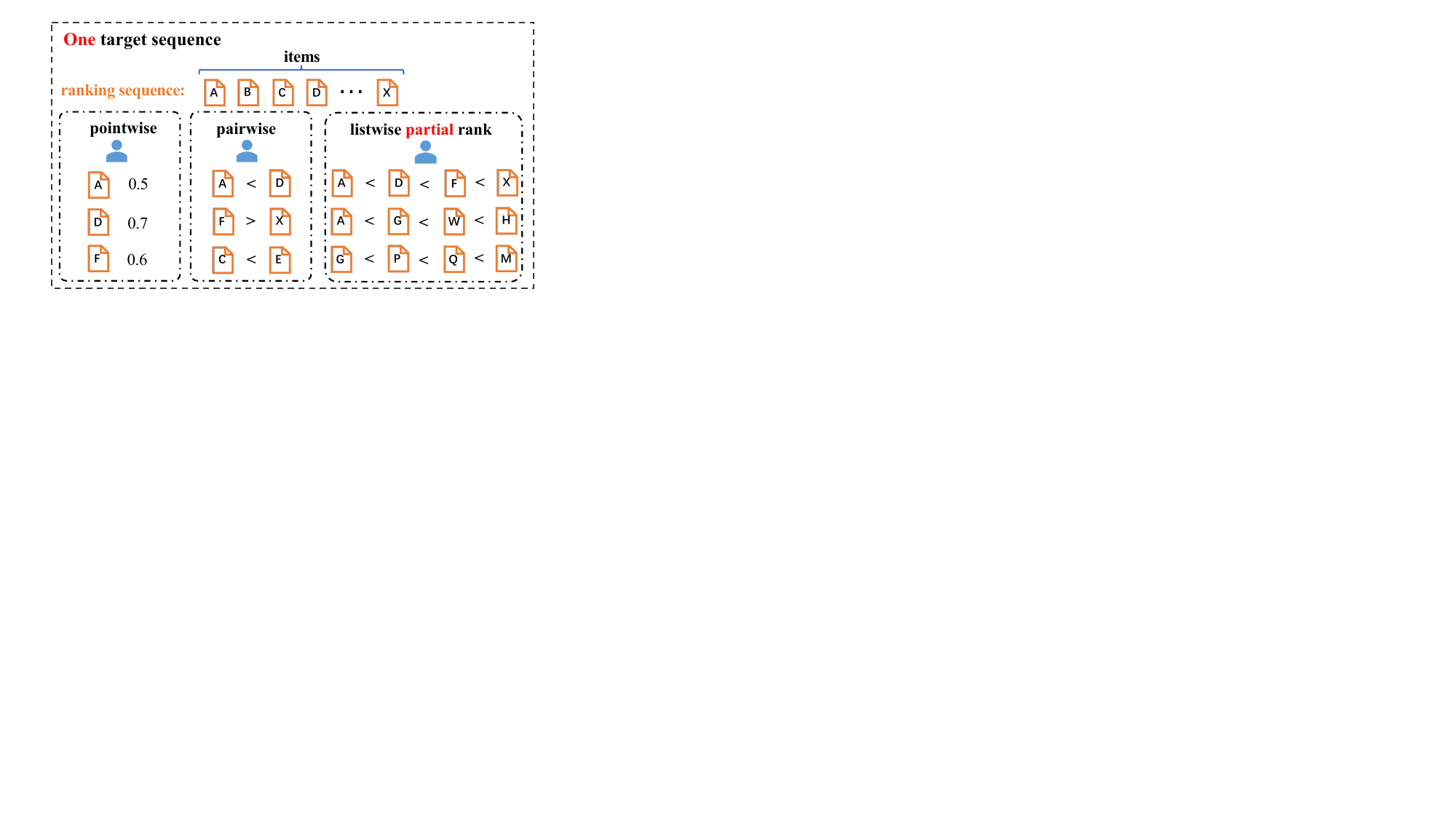}
  }
    \end{minipage}
  
  \caption{Comparison between different rank aggregation paradigms. In this figure, $A\sim Z$ or $A_i\sim D_i$ are items to be ranked, and their ground-truth ranks are determined according to the alphabetical order. (a): In previous rank aggregation settings, the ground-truth rank of a single long sequence should be determined.
  In pointwise methods, only ratings for the assigned items are provided by each annotator, while in pairwise methods, the relative rankings within pairs of items are provided. For the listwise partial rank aggregation task, each annotator receives a subset of items and provides the corresponding ranks over them. (b): In the context of listwise full rank aggregation, there are multiple short sequences, and the items within each sequence remain to be ranked. During the annotation process, each annotator is assigned several sequences and is required to annotate complete ranks for the items in each sequence.}
  \label{fig: comparison}
\end{figure}

Existing rank aggregation methods can be roughly classified into three categories according to different forms of annotations: pointwise, pairwise, and listwise. Fig.~\ref{fig: comparison} illustrates the similarities and differences between these forms of annotations. 
Under the setting of pointwise methods, each sub-task contains only one item, and the annotator assigns scores independently to each item without access to the information of other items. The aggregation methods will then derive one ranking sequence based on these score annotations~\cite{boada_count}.
In the pairwise setting, the sub-task contains one pair of items, and the annotator determines the relative ranking between two items. The aggregation methods use the results of compared pairs to form the final rank~\cite{greed_order}.
In the listwise setting, the sub-task contains multiple items, which is a subset of all the items. The annotator is required to provide the rank of this subset, and the aggregation methods aggregate ranks of all subsets into a final rank~\cite{crowd_agg_alg}. These types of methods find applications in recommendation systems~\cite{recommendation_systems}, information retrieval~\cite{information_retrieval}, bioinformatics~\cite{Bioinformatics}, etc.
Obviously, although the forms of annotations are different, all these types of methods aim to derive the ground-truth rank of \textbf{one} target ranking sequence.

However, the above forms of annotation may not cover all the rank aggregation problems. As shown in Fig.~\ref{fig: comparison}\,(b), there also exists another scenario in which we need to deduce the ranks of multiple target sequences based on full rank annotations. Specifically, there are \textbf{multiple short sequences} required to be ranked, with each sub-task containing all items of one sequence. The annotator is required to give a full rank over each assigned sequence. Under this setting, different annotators are possible to label the same target sequence, leading to redundant annotations. The aggregation method is hereby expected to infer the ground-truth rank of each sequence simultaneously. Such a setting can be applied to tasks such as model quality assessment~\cite{survey_evaluation_LLMs} and reinforcement learning with human feedback (RLHF)~\cite{RLHF_application}. To be specific, in the model quality assessment task, the outputs of several models are evaluated in each round, and these outputs form a single sequence (like $A_1, B_1, C_1, D_1$ in Fig.~\ref{fig: comparison}\,(b)). Similarly, in RLHF, the outputs of a large language model are sampled, forming a sequence to be ranked by human annotators. Table~\ref{tab: different_paradigms} further outlines the differences between various aggregation tasks and their applications, where it is highlighted that listwise full rank aggregation aims to handle multiple short sequences with full rank annotations.

\setlength{\tabcolsep}{3.5pt}
 \begin{table}[!t]
 \centering
    \tiny
  \caption{Properties of various rank aggregation tasks and their applications.}
  \begin{tabularx}{\textwidth}{lcccc}
    \toprule
    \multicolumn{1}{c}{Scenarios} & Multiple sequences? & Short sequences? & \makecell{Partial annotations \\for each sequence?} & \multicolumn{1}{c}{Typical applications}\\
    \midrule
    pointwise rank aggregation & $\times$ & $\times$ & $\surd$ & \multirow{3}*{ \makecell{1) recommendation systems~\cite{recommendation_systems}\\2) information retrieval~\cite{information_retrieval} \\3) bioinformatics~\cite{Bioinformatics}}}\\
    pairwise rank aggregation & $\times$ & $\times$ & $\surd$ & ~\\
    listwise partial rank aggregation & $\times$ & $\times$ & $\surd$ & ~\\
    \textbf{listwise full rank aggregation} & $\boldsymbol{\surd}$ & $\boldsymbol{\surd}$ & $\boldsymbol{\times}$ & \makecell{
    \textbf{1) model quality assessment~\cite{survey_evaluation_LLMs}}\\
    \textbf{2) reinforcement learning with human feedback~\cite{RLHF_application}}
    }\\
    \bottomrule
  \end{tabularx}
  \label{tab: different_paradigms}
\end{table}

Various methods have been proposed to tackle pointwise, pairwise, and listwise rank aggregation problems. The pointwise methods often minimize a specific distance between the collected scores and the aggregated ones~\cite{median_rank}, while the pairwise methods usually model the pairwise probability between items~\cite{condorcet_fuse}. Besides, the listwise partial rank aggregation methods often convert the aggregation task to the pairwise one with the consideration of uncertainty~\cite{Stagg_alg}.  Although some of these methods can be adapted to full rank annotations by solving each problem independently, they fail to characterize the relationship between different problems and cannot capture the global information in full rank annotations. Therefore, in this paper, we present a specialized study for listwise full rank aggregation to fill the gaps in research and practice.

The full rank aggregation task is intrinsically difficult since different sequences have different levels of difficulty, and the ability of annotators can be different. Therefore, explicit modeling of the ability of annotators and the difficulty of problems is desired, which is usually a missing concern in previous methods. Specifically, the ability of an annotator is supposed to reflect the extent of uncertainty associated with the rank position of each candidate item. Therefore, the probability that the annotator flips every pair of candidate items should be considered. Meanwhile, the difficulty of problems can also be represented by multiple probabilities, measuring whether each pair of items is prone to confusion. In addition, from a global point of view, in the final rank list, the nearby items are generally much more likely to be mispositioned than those with a more significant gap in position. Therefore, the ability of annotators and the difficulty of problems should be dynamically adjusted according to the gap in position.

To this end, we propose a \underline{\textbf{L}}istwise rank \underline{\textbf{A}}ggregation method in \underline{\textbf{C}}rowdsourcing~(``LAC" for short hereinafter), to deal with the listwise input in crowdsourcing. To the best of our knowledge, this method represents the first endeavor specifically addressing the task of full rank aggregation in crowdsourcing. In our method, for the ability of annotators and the difficulty of problems, two sets of confusion matrices are employed to estimate the degree of confusion for each pair of items. Then, the distance between positions is carefully defined to integrate the relative positional information between two items. Considering the unknown latent variables of the ability of annotators, the difficulty of problems, and the true rank, we derive the log-likelihood of the observations and maximize it with the Expectation Maximization~(EM) method. Specifically, in the E-step, based on the estimated values of latent variables derived in the previous step, we obtain the conditional expectation of observations over the underlying rank distribution. In the M-step, the latent variables and the ground-truth ranks are estimated respectively via maximizing the expectation calculated in the E-step.  

In experiments, we propose to synthesize ranking datasets with the consideration of five essential factors, including number of problems, length of rank, number of annotators,  ability of annotators, and the annotation ratio, so that the performance in different scenarios can be investigated. Furthermore, a real-world dataset with a total of 5,981 listwise ranking annotations from 25 annotators is collected to examine the performance of LAC.\par
Our contributions can be summarized in three folds: 
\begin{itemize}
\item We focus on the underexplored problem of listwise full rank aggregation in crowdsourcing and propose LAC as a new rank aggregation method.
\item Algorithmically, the ability of annotators, the difficulty of problems, and the ground-truth ranks are respectively modeled as latent variables, and we propose to use the EM method to deduce their optimal values iteratively.
\item Experimentally, we simulate synthetic datasets with a comprehensive consideration of five essential factors to explore the performance of LAC. 
We also present, for the first time, a real-world dataset under the listwise full rank aggregation setting to benchmark previous methods. Experimental results on both synthetic and real-world datasets demonstrate the superiority of LAC over existing methods in most scenarios.    
\end{itemize}

\section{Related Work}
\noindent In this section, we introduce some related works, including the background of crowdsourcing and the existing rank aggregation methods. 
\subsection{Crowdsourcing}
There are three critical concerns in crowdsourcing, \ie, cost control~\cite{zheng_survey}, latency control~\cite{zeng2018latency}, and quality control~\cite{allahbakhsh2013quality, tkdd_a_lightweight_effective_crowdsourcing_2024}. \par
\textbf{Cost control}. Crowdsourcing may be expensive when dealing with a large number of tasks and instances. In order to alleviate this issue, several cost-control techniques have been proposed. These techniques include removing unnecessary tasks and picking up valuable tasks~(task pruning~\cite{wang2012crowder}), ranking and prioritizing valuable tasks (task selection~\cite{guo2012so}), deducing answers for the candidate tasks based on the feedback data~(answer deduction~\cite{wang2013leveraging}), and sampling tasks based on some specific criteria for crowdsourcing.
\par
\textbf{Latency control}. Crowdsourcing for answering tasks may suffer from excessive latency due to the unavailability of annotators, the difficulty of tasks, and insufficient appeal to annotators. Therefore, latency control is in need. Two representative models for latency control have been proposed, namely the round model~\cite{sarma2014crowd} and the statistical model~\cite{yan2010crowdsearch}. Here, the round model arranges tasks to be published in many rounds and models the overall latency as the number of rounds. In contrast, the statistical model uses feedback data to build models that capture the arrival and completion times of annotators, allowing better prediction and adjustment for expected latency.
\par
\textbf{Quality control}. Crowdsourcing may produce low-quality or even incorrect answers due to annotators' varying levels of expertise. Therefore, quality control is crucial. 
 The ability of annotators can be modeled and controlled through several methods, including eliminating low-quality annotators~\cite{ipeirotis2010quality}, aggregating answers from multiple annotators~\cite{cao2012whom}, and assigning tasks to appropriate annotators based on their skills and experience~\cite{zhao2015crowd}. How to infer the true labels from multiple noisy labels is a critical problem in quality control. The most seminal work is DS~\cite{DS}, which uses a confusion matrix to represent the quality of the crowdsourcing annotator. Then several works are derived from the DS algorithm directly. For example, \cite{DS_simplify_version} simplifies the parameters of the annotators, \cite{DS_constrain_version} constrains the annotator in some perspectives, and \cite{DS_parameters} optimizes the initial settings. In addition, another stream of approaches models the quality of an annotator using fewer parameters~\cite{crowdsourcing_parameters_1} while adding auxiliary parameters to model the annotator's property, such as bias ~\cite{bias_method_1} and intention~\cite{crowdsourcing_parameters_1}. All of them are based on probabilistic models and can be solved via an EM algorithm with gradient descent. Some of them~\cite{crowdsourcing_parameters_1, DS} can be applied to multi-class scenarios, while others~\cite{DS_simplify_version, crowdsourcing_parameters_3} are only suitable for binary classification problems. Unfortunately, none of these techniques can handle listwise full rank aggragation problem. Next, we introduce some methods to aggregate multiple ranks.

\subsection{Rank Aggregation Methods} 
According to various input forms, the existing ranking methods can be roughly classified into three categories, namely pointwise, pairwise, and listwise methods. 
\par
\textbf{Pointwise}. Two well-known pointwise methods, Borda Count~\cite{boada_count} and Median Rank~\cite{median_rank}, are often adopted to obtain suitable rankings. Specifically, Borda Count minimizes the average Spearman Rank Coefficient, and Median Rank minimizes the average Spearman Footrule Distance between the true rank and each input.
\par
\textbf{Pairwise}. Pairwise rank aggregation methods organize their ranking inputs in pairs and optimize the objective function or ranking functions accordingly~\cite{tkdd_rank_from_crowdsourced_pairwise_2020}. For example, BradleyTerry~\cite{BradleyTerry} defines the pairwise probability based on the BradleyTerry model and then optimizes the likelihood function by gradient descent.
Afterwards, GreedyOrder~\cite{greed_order} focuses on minimizing the cost of pairwise disagreement in a tournament to infer the true rank. In contrast, CondorcetFuse~\cite{condorcet_fuse} builds a Condorcet Graph by majority voting and obtains a Hamiltonian path from the graph by QuickSort, and SVP~\cite{SVP_nuclear} minimizes the nuclear norm of a pairwise preference matrix by rank-2 factorization. However, pairwise methods usually cannot capture the global information over items.
\par
\textbf{Listwise}. Unlike the pointwise and pairwise methods, listwise rank aggregation methods treat the ranking inputs in a listwise way to emphasize the importance of positions. Previous listwise methods usually aim at the aggregation of partial ranks~\cite{tkdd_listwise_learning_2016}. The typical methods include Plackett-Luce~\cite{pluckett_luce}, St.Agg~\cite{Stagg_alg}, and CrowdAgg~\cite{crowd_agg_alg}. Among them, Plackett-Luce extends the Plackett-Luce model~\cite{BT_base_model} and defines the similarity of ranks based on generative probability. St.Agg incorporates uncertainty into the aggregation process and introduces a prior rank distribution. Besides, CrowdAgg further takes the quality of annotators into consideration. Unfortunately, the above methods primarily focus on partial ranks, where only part of a single long sequence is required to be ranked by each annotator (see Fig.~\ref{fig: comparison}). Meanwhile, the study directly dealing with full ranks over the items is basically in the blank. Since such a setting is of utmost significance in real-world applications, we thereby undertake meticulous studies on the problem of listwise full rank aggregation.

\section{The Proposed Method}
\subsection{Preliminary}
In this section, we formalize the listwise full rank aggregation task in crowdsourcing. Suppose that there are totally $I$ problems\,/\,sequences and $J$ crowdsourced annotators. The dataset with ground-truth ranks is denoted by $\mathcal{S}=\{(x_i,y_i)\}_{i=1}^{I}$, where $x_i$ represents the $i$-th problem, and the underlying true rank is $y_i$. Each problem posted on the crowdsourcing platform is associated with $R$ items, and the selected annotators provide ranks over these items based on their knowledge. The ground-truth rank $y_i$ and the ranks provided by the selected annotators are permutations of the items in $x_i$. All the true ranks of the entire dataset are denoted by $\mathbf{Y}=[y_1, \cdots, y_I]$. Let $E_j$ be the $j$-th annotator. The rank given by $E_j$ for the problem $x_i$ is denoted by $l_{ij}$, and all ranks provided for this problem are denoted by $l_i$. Then the observed dataset is denoted by $\mathcal{L}=\{(x_i,l_i)\}_{i=1}^{I}$. Obviously, the resulting annotations for each problem consist of noisy repeated ranks. Therefore, our task is to deduce the ground-truth rank $y_i$ for each problem $x_i$ based on the noisy ranks $l_i$. To this end, we use $d$ and $k$ to represent lists and define $\tau(k,d_r)$ as the index of $d_r$ in list $k$, where $d_r$ indicates the item in position $r$ of list $d$. The main mathematical notations that will be later used for algorithm description are listed in Table~\ref{tab: mathematical_notations}.

 \begin{table}[!t]
  \caption{Summary of main mathematical notations}
  \begin{tabular}{cp{0.7\linewidth}}
    \toprule
    \multicolumn{1}{c}{Notation} & \multicolumn{1}{c}{Interpretation}\\
    \midrule
    $I$& the number of problems\,/\,sequences. \\
    $J$&  the number of annotators. \\
    $R$&  the number of items in a single problem. \\
    $x_i$& the $i$-th problem. \\
    $y_i$& the unknown ground-truth rank of problem $x_i$. \\
    $E_j$&  the $j$-th crowdsourced annotator. \\
    $l_{ij}$ & the rank of problem $x_i$ given by annotator $E_j$. \\
    $\Pi^{(j)}$ & the confusion matrix that models the ability of the annotator $E_j$. \\
    $\Delta^{(i)}$ & the confusion matrix that models the difficulty of the problem $x_i$. \\
    $\theta_k$& the likelihood for the true rank $k$. \\
    $\pi_{k_rd_r}^{(j)}$ & the probability of item $k_r$ being confused with $d_r$ by annotator $E_j$. \\
    $\delta_{k_rd_r}^{(i)}$ & the probability of confusion between item $k_r$ and $d_r$ in problem $x_i$. \\
    $d,\; k$ & a ranked list. \\
    $\mathbf{Y}$ & the list containing all the ground-truth ranks.\\
    $\mathcal{L}$ & the noisily ranked dataset collected from the crowdsourcing platform.\\
    $\Psi$ &  the set $\{\boldsymbol{\theta}, \Pi, \Delta\}$.\\
    \bottomrule
  \end{tabular}
  \label{tab: mathematical_notations}
\end{table}

\subsection{Listwise Rank Aggregation in Crowdsourcing}
\label{likelihood part}
    
\begin{figure}[t]
    \centering
    \includegraphics[width=0.4\linewidth]{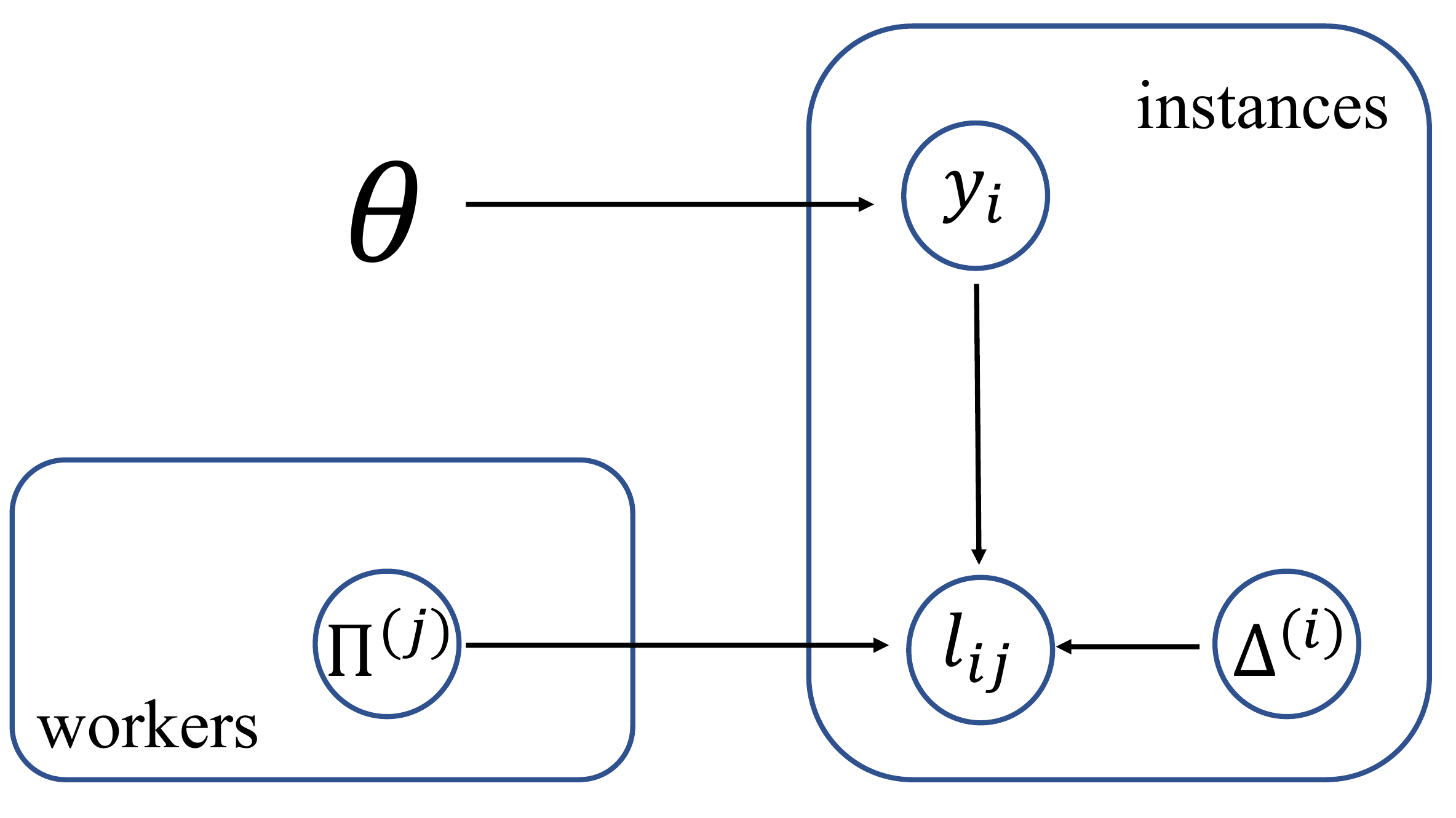}
    \caption{Probabilistic graphical model representation of the proposed LAC method.}
    \label{prob_graph}
\end{figure}

    To tackle the task introduced above, we propose a novel probabilistic generative model termed LAC. Unlike previous methods that may only consider the quality of annotators~\cite{crowd_agg_alg}, we also incorporate the difficulty of problems into our model, which can further capture the intrinsic property of the aggregation task. Intuitively, it is reasonable that the crowdsourced annotator with stronger ability can perform better when provided with easier problems and vice versa. Therefore, we introduce a confusion matrix $\Pi^{(j)}\in[0,1]^{R\times R}$ and $\Delta^{(i)}\in[0,1]^{R\times R}$ for the annotator $E_j$ and the problem $x_i$, respectively. Here $\Pi^{(j)}$ explicitly characterizes the ability of the $j$-th annotator, namely, the $(k_r,d_r)$-th element $\pi_{k_rd_r}^{(j)}$ denotes the probability of $k_r$ being confused with $d_r$ by $E_j$. Similarly, $\Delta^{(i)}$ explicitly characterizes the difficulty of the $i$-th problem, namely, the $(k_r,d_r)$-th element $\delta_{k_rd_r}^{(i)}$ denotes the probability of $k_r$ being confused with $d_r$ in $x_i$. To be clear, we take $R=3$ for example. Suppose that the ground-truth rank of the problem $x_i$ is given as $k = A \prec B \prec C$, and $E_j$'s annotation is given as $d = B \prec A \prec C$. When $r=1$, we can get $k_1=A$ and $d_1=B$, and then we have $\pi_{k_1d_1}^{(j)}=\pi_{AB}^{(j)}$, which represents the probability of $A$ being confused with $B$ by annotator $E_j$, and $\delta_{k_1d_1}^{(i)}=\delta_{AB}^{(i)}$, which represents the probability of $A$ being confused with $B$ in problem $x_i$. The examples of $\Pi^{(j)}$ and $\Delta^{(i)}$ are given by
    \begin{equation}
    \renewcommand{\arraystretch}{1.3}
    \label{pi_j_matrix}
    \Pi^{(j)}=\begin{bmatrix}
     \pi_{AA}^{(j)} & \pi_{AB}^{(j)} &  \pi_{AC}^{(j)}\\
     \pi_{BA}^{{ (j)}} & \pi_{BB}^{(j)} &  \pi_{BC}^{(j)}\\
     \pi_{CA}^{{ (j)}} & \pi_{CB}^{(j)} &  \pi_{CC}^{(j)}\\
    \end{bmatrix}_{3 \times 3},   
    \end{equation}
    and
    \begin{equation}
    \renewcommand{\arraystretch}{1.3}
     \Delta^{(i)}=\begin{bmatrix}
    \delta_{AA}^{(i)} & \delta_{AB}^{(i)} &  \delta_{AC}^{(i)}\\
    \delta_{BA}^{(i)} & \delta_{BB}^{(i)} &  \delta_{BC}^{(i)}\\
    \delta_{CA}^{(i)} & \delta_{CB}^{(i)} &  \delta_{CC}^{(i)}\\
    \end{bmatrix}_{3 \times 3},
    \end{equation}
    respectively. The probabilistic graphical model is illustrated in Fig.~\ref{prob_graph}, and there are three latent variables referred to as $\Psi=\{ \boldsymbol{\theta}, \Pi, \Delta \}$. Therefore, our target is transformed to deriving the most reasonable value of $\Psi$ to maximize the likelihood of all observations $\{l_i\}_{i=1}^I$ and inferring the possible ranks based on the optimal $\Psi$.
    
\textbf{Likelihood of the observed dataset.}
    Suppose that the ground-truth rank of each problem is independently drawn from a multinomial distribution with parameter $\boldsymbol{\theta}= [ \theta_{1}, \cdots, \theta_{K}]$, where $\sum_{k=1}^{K}\theta_{k}=1$, and $K$ is the total number of permutations over all items. Here, $\theta_{k}$ is the prior probability, defined as
    \begin{equation}
        \theta_{k}=p(y_i=k|\boldsymbol{\theta}).
    \end{equation}  
    Since each problem is independently annotated, the likelihood of the observed dataset $\mathcal{L}$ can be factorized as
    \begin{equation}\label{likehood_whole_data}
        \begin{aligned}
            P(\mathcal{L}|\Psi)=\prod_{i=1}^{I}P({l_i|\Psi}),
        \end{aligned}
    \end{equation}
    which is governed by the parameter set $\Psi=\{\boldsymbol{\theta}, \Pi, \Delta \}$. Subsequently, the likelihood of the annotation $l_i$ can be derived as 
    \begin{equation}\label{li_likelihood}
        \begin{aligned}
            P(l_i|\Psi)=\sum_{k=list_{first}}^{K=list_{last}}\underbrace{P(y_i=k|\boldsymbol{\theta})}_{\text{the prior } \theta_k} \underbrace{P(l_i|y_i=k,\Pi,\Delta^{(i)})}_{\text{the posterior probability}}.
        \end{aligned}
    \end{equation}
    Since our input is in listwise form, the global ranking information can be readily obtained. It is crucial to utilize this information to calculate the distance between the annotated rank and the possible true rank. In view of this point, we calculate the likelihood when $k$ is wrongly ranked as $d$ in a position-wise way, where we multiply the reciprocal of the distance for penalizing those ranks with further distance. Therefore, the posterior probability is defined as
    \begin{equation}
    \label{li_likelihood_y_i}
            P(l_i|y_i=k,\Pi,\Delta^{(i)})=\prod_{j=1}^{J}\prod_{d=list_{first}}^{K=list_{last}}\big(\underbrace{\prod_{r=1}^{R}(\frac{\pi_{k_rd_r}^{(j)}\delta_{k_rd_r}^{(i)}}{ d(k_r,d_r) }) }_{\mathcal{F}(\Pi^{(j)},\Delta^{(i)})}\big)^{\mathbb{I}(l_{ij}=d)},
    \end{equation}
    where $d(k_r,d_r)=|\tau(k,d_r)-r|+1$. For brevity, in the following derivation, we denote 
    \begin{equation}
    \begin{aligned}
        \mathcal{F}(\Pi^{(j)},\Delta^{(i)})=\prod_{r=1}^{R}\big(\frac{\pi_{k_rd_r}^{(j)}\delta_{k_rd_r}^{(i)}}{ d(k_r,d_r) }\big).
        \end{aligned}
    \end{equation}
    
    To enhance clarity, here we introduce an example to show the rationale of the position-wise distance. Let $R=3$, and we suppose that the ground-truth rank is $A \prec B \prec C$. Assume that the first annotator gives the rank $k=B \prec A \prec C $, and the second one gives the rank $d=C \prec B \prec A$. For simplicity, we only take the first position of each rank as an example. For the first annotator, the distance in the first position is 1 (because the difference of indices between $B$ and $A$ is 1). But for the second annotator, the distance in the first position is 2 (because the difference of indices between $C$ and $A$ is 2). It is evident that for the first position, $k$ is better than $d$. Therefore, it is reasonable to take the distance as a penalty.
    
    Subsequently, the log-likelihood of the noisy repeated listwise rank can be represented as
    \begin{equation}\label{dataset_likelihood}
        \begin{aligned}
            \sum_{i=1}^{I}\ln{\left(\sum_{k=list_{first}}^{K=list_{last}}\theta_{k}P(l_i|y_i=k,\Pi,\Delta^{(i)})  \right)},
        \end{aligned}
    \end{equation}
    where $P(l_i|y_i=k,\Pi,\Delta^{(i)})$ is defined in Eq.~\eqref{li_likelihood_y_i}.
    
    \subsection{Optimization with EM Algorithm}\label{E_step}
    To effectively seek the latent variables that maximize the log-likelihood defined in Eq.~\eqref{dataset_likelihood}, we employ the prevalent Expectation Maximization~(EM) algorithm~\cite{PRML}, which estimates the $\boldsymbol{\theta}$, the element of $\Pi^{(j)}$, and $\Delta^{(i)}$ iteratively.\\

    \textbf{E-step.} We first construct an expectation function of the log-likelihood of the dataset on latent variable $\mathbf{Y}$ as follows:
    \begin{equation}
        \begin{aligned}
          Q(\Psi, \Psi^{old})=\mathbb{E}_{\mathbf{Y}|L, \Psi^{old}}[ \ln P(L,\mathbf{Y}|\Psi) ],
        \end{aligned}
    \end{equation}
    where $\Psi^{old}$ is the value of $\Psi$ in the previous step.
    
    Afterwards, we calculate the expectation over the possible latent variable $y_i$, namely
    \begin{equation}\label{E_step_equation}
        \begin{aligned}
            \mathbb{E}_{y_i}[\mathbb{I}(y_i=k)]&= P(y_i=k|L,\Psi) \propto P(L|y_i=k,\Psi)P(y_i=k|\Psi)  \\
            &=\theta_k\prod_{j=1}^{J}\prod_{d=list_{first}}^{K=list_{last}}(\mathcal{F}(\Pi^{(j)},\Delta^{(i)}))^{\mathbb{I}(l_{ij}=d)}.
        \end{aligned}
    \end{equation}
	
    \textbf{M-step}. The parameters $\boldsymbol{\theta}$, $\Pi$, and $\Delta$ are updated to maximize the function $Q$. 
    In the first step, we expand the function of $Q$ using Bayes' Theorem, namely
	\begin{equation}\label{equation_Q}
	\begin{aligned}
	    \mathbb{E}_{\mathbf{Y}|L, \Psi^{old}} \left[  \ln P(L,\mathbf{Y}|\Psi) \right]
	    = & \sum_{i=1}^{I}\mathbb{E}_{y_i}\left[ \ln P(l_i,y_i|\boldsymbol{\theta},\Pi,\Delta^{(i)}) \right]\\ 
	    = & \sum_{i=1}^{I}\mathbb{E}_{y_i}\left[ \ln \left( P(l_i|y_i, \boldsymbol{\theta},\Pi,\Delta^{(i)})P(y_i| \boldsymbol{\theta})  \right)  \right].
	    \end{aligned}
	\end{equation}
    Because the prior probability $P(y_i|\boldsymbol{\theta})$ is a constant w.r.t. $\Psi$ in derivatives, we only need to maximize the term $\sum_{i=1}^{I}\mathbb{E}_{y_i}\left[ \ln P(l_i|y_i, \boldsymbol{\theta},\Pi, \Delta^{(i)}) \right]$. Consequently, Eq.~\eqref{equation_Q} can be reformulated as
	\begin{equation}\label{m_step_likihood}
	\begin{aligned}
	    &\mathbb{E}_{\mathbf{Y}|L, \Psi^{old}}\left[  \ln P(L,\mathbf{Y}|\Psi) \right] = \sum_{i=1}^{I}\sum_{k=list_{first}}^{K=list_{last}} \mathbb{E}\left[ \mathbb{I}(y_i=k) \right]\\
	    &\cdot \left[ \ln \theta_k + \sum_{j=1}^{J}\sum_{d=list_{first}}^{K=list_{last}}\mathbb{I} (l_{ij}=d)\ln (\mathcal{F}(\Pi^{(j)},\Delta^{(i)})) \right] .
	    \end{aligned}
	\end{equation}

    The parameters $\boldsymbol{\theta}$, $\Pi$, and $\Delta$ are updated to maximize the function $Q$. 
    \paragraph{Update on $\theta_k$} Here, we use the Lagrange multiplier to find the optimal $\theta_k$ in Eq.~\eqref{m_step_likihood}, and we construct the function
	\begin{equation}
	    F_1(\theta_k)=\sum_{i=1}^{I}\mathbb{E}[ \ln P(l_i|y_i, \boldsymbol{\theta}, \Pi, \Delta^{(i)} ) ] +\lambda\left( \sum_{k=1}^{K}\theta_k-1  \right).
	\end{equation}
    Then, we set $\frac{\partial F_1}{\partial \theta_k} = 0$, \ie,
    
    \begin{equation}\label{before_equation}
        \frac{\partial F_1}{\partial \theta_k}=\frac{\sum_{i=1}^{I}\mathbb{E}[\mathbb I(y_i=k)]}{\theta_k}+\lambda=0.\\
    \end{equation}	
    By applying the sum-up-to-one condition, namely $\sum_{k=1}^{K}\theta_k=1$,  we can obtain $\lambda=-I$. Subsequently, plug it into Eq.~\eqref{before_equation}, we can obtain the estimation of $\theta_k$, which is given by
    \begin{equation}
    \label{p_k_equation}
        \hat{\theta}_k=\frac{\sum_{i=1}^{I}\mathbb{E}[ \mathbb I(y_i=k) ] }{I}.
    \end{equation}
    
    \paragraph{update on $\pi_{k_rd_r}^{(j)}$}
    By using the Lagrange multiplier to optimize $\pi_{k_rd_r}^{(j)}$ in Eq.~\eqref{m_step_likihood}, we construct the function
    \begin{equation}
        F_2=\sum_{i=1}^{I}[\ln P(l_i|y_i, \boldsymbol{\theta}, \Pi, \Delta^{(i)})]+\alpha \left( \sum_{r=1}^{R}\pi_{k_rd_r}^{(j)}-1 \right),
    \end{equation}
    and then we set the partial derivative $\frac{\partial F_2}{\partial \pi_{k_rd_r}^{(j)}}$ to zero, namely,
	\begin{equation}\label{pi_equation_before}
	    \frac{\partial F_2}{\partial \pi_{k_rd_r}^{(j)}}=\frac{\sum_{i=1}^{I}\{ \mathbb{E} [ \mathbb{I} (y_i=k) ] \mathbb{I}(l_{ij}=d) \}}{\pi_{k_rd_r}^{(j)}}+\alpha=0.
	\end{equation}
	With the sum-up-to-one condition ($\sum_{r=1}^{R} \pi_{k_rd_r}^{(j)}=1$) , we have $\alpha=-\sum_{i=1}^{I}\{\mathbb{E}[\mathbb{I}(y_i=k) ] \}$. By plugging it into Eq.~\eqref{pi_equation_before}, we can get the optimal $\pi_{d_rk_r}^{(j)}$, which is formulated as
	\begin{equation}
        \label{pi_kd}
        \begin{aligned}
         \hat{\pi}_{k_rd_r}^{(j)}=\frac{\sum_{i=1}^{I}\{ \mathbb{E} [ \mathbb{I} (y_i=k) ] \mathbb{I}(l_{ij}=d) \}}{\sum_{r=1}^{R} \sum_{i=1}^{I}\{ \mathbb{E} [ \mathbb{I} (y_i=k) ]\mathbb{I}(l_{ij}=d) \}}.
        \end{aligned}
    \end{equation}
    
    \paragraph{update on $\delta_{k_rd_r}^{(i)}$}
    By using the Lagrange multiplier to optimize $\delta_{k_rd_r}^{(i)}$ in Eq.~\eqref{m_step_likihood}, we construct the function
    \begin{equation}
        F_3=\sum_{i=1}^{I}[\ln P(l_i|y_i, \boldsymbol{\theta}, \Pi, \Delta^{(i)})]+\beta \left( \sum_{r=1}^{R}\delta_{k_rd_r}^{(i)}-1 \right),
    \end{equation}
    and then we set the partial derivative $\frac{\partial F_3}{\partial \delta_{k_rd_r}^{(i)}}$ to zero. Applying the sum-up-to-one condition ($\sum_{r=1}^{R}\delta_{k_rd_r}^{(i)}=1$), we have
	\begin{equation}\label{delta_kd_before}
	    \frac{\partial F_3}{\partial \delta_{k_rd_r}^{(i)}}=\frac{ \mathbb{E} [ \mathbb{I} (y_i=k) ]\sum_{j=1}^{J} \mathbb{I}(l_{ij}=d) }{\delta_{k_rd_r}^{(i)}}+\beta=0.
	\end{equation}
	With this condition, we have $\beta=-J\{\mathbb{E}[\mathbb{I}(y_i=k) ] \}$. By plugging it into Eq.~\eqref{delta_kd_before}, we obtain the estimation of $\delta_{d_rk_r}^{(i)}$, given by 
	\begin{equation}
        \label{delta_kd}
        \begin{aligned}
         \hat{\delta}_{k_rd_r}^{(i)}=\frac{ \sum_{j=1}^{J} \mathbb{I}(l_{ij}=d)\mathbb{E} [ \mathbb{I} (y_i=k) ] }{J\{ \mathbb{E} [ \mathbb{I} (y_i=k) ] \}}. 
        \end{aligned}
    \end{equation}

\renewcommand\arraystretch{0.8}	
\begin{algorithm}[t]
\small
	\caption{The overall algorithm of LAC.}
	\label{algorithm_pipeline}
	\begin{algorithmic}[1]
	    \STATE{\bf Input:} Listwise full rank dataset $\mathcal{L}$, number of problems $I$, number of annotators $J$.  
		\FOR {$i=0 \cdots I$}
 	        \STATE initialize the difficulty matrix $\Delta^{(i)}$.
 	      \ENDFOR
 	    \FOR {$j=0 \cdots J$}
 	        \STATE initialize the ability matrix $\Pi^{(j)}$.
 	    \ENDFOR
 	    \WHILE{ not converge }
 	         \STATE E-step:  
 	         \STATE    \quad \quad \quad   calculate the expectation by Eq.~\eqref{E_step_equation}.
 	         \STATE M-step: \\
                \STATE    \quad  \quad \quad  update $\theta_k$ via Eq.~\eqref{p_k_equation}.
                \STATE \quad  \quad \quad update $\pi_{k_rd_r}$ via Eq.~\eqref{pi_kd}.
                \STATE \quad  \quad \quad update $\delta_{k_rd_r}$ via Eq.~\eqref{delta_kd}.
 	    \ENDWHILE
 	    \STATE For each problem $x_i$, calculate $\hat{y}_i$ by Eq.~\eqref{calculate_y_i}.
	    \STATE \textbf{Output:} the inferred ranks $\{\hat{y}_i\}_{i=1}^I$.\\
	\end{algorithmic}
\end{algorithm}

    \textbf{After convergence.}
    Based on Eq.~\eqref{E_step_equation}, the inferred true rank of the $i$-th problem can be determined as
	\begin{equation}\label{calculate_y_i}
	    \begin{aligned}
	        \hat{y}_i=\arg \max_k \mathbb{E}[\mathbb{I}(y_i=k)].
	    \end{aligned}
	\end{equation}
The detailed steps of the proposed LAC method are summarized in Algorithm~\ref{algorithm_pipeline}. 

\subsection{Complexity Analysis}
In this section, we analyze the computational complexity of the proposed LAC method.

Since our method can be divided into the E-step and the M-step, we first analyze the complexity of the E-step in Algorithm~\ref{algorithm_pipeline}. We use $\eta$ to denote the annotation ratio, \ie, the ratio of annotators required to provide ranks for each problem. Therefore, for each problem, $\eta\cdot J$ annotators give the full rank, where $J$ is the number of annotators, and thus the complexity of E-step in Eq.~\eqref{E_step_equation} is $\mathcal{O}(I\cdot R!\cdot \eta \cdot J)$. In the M-step, the calculation of $\theta_k$ requires $\mathcal{O}(R!\cdot I)$ computations, the calculation of $\pi_{k_rd_r}$ requires $\mathcal{O}(R^3\cdot J\cdot I)$ computations, and the calculation of $\delta_{k_rd_r}$ requires $\mathcal{O}(I\cdot R^2\cdot J)$ computations. Therefore, taking all the above results into consideration and suppose that the total iterations required are $T$, our LAC algorithm takes $\mathcal{O}(T\cdot( I\cdot R!\cdot \eta \cdot J+R!\cdot I+R^3\cdot J\cdot I))$ complexity. From the above analysis, we conclude that our algorithm applies for some cases with a moderate number of problems $I$ and a relatively small length $R$ for each problem. Here, we take the model quality assessment task as an example. There are often few models to be compared (usually three to seven)~\cite{RLHF_application}, rendering the complexity of our LAC acceptable.



\section{Experiments}
\noindent In this section, we conduct intensive experiments on both synthetic and real-world datasets to demonstrate the superiority of the proposed method. The implementation code of LAC can be found at \url{https://anonymous.4open.science/r/LAC-B871}.

\subsection{Experimental Setting}
The target of rank aggregation is to correctly infer the truth at each position. Therefore, we utilize a position-wise accuracy metric to evaluate the overall performance, defined as
\begin{equation}\label{acc_equation}
    \begin{aligned}
        acc=\frac{\#\,correctly\,predicted\, positions}{\#\,  positions\, in\,all\,problems },
    \end{aligned}
\end{equation}
where each correctly predicted position represents an aggregated result identical to the true rank in some positions.

 To validate the effectiveness of the proposed LAC method, we choose five baseline methods for comparison. Specifically, the pairwise methods adopted are BradleyTerry~\cite{BradleyTerry}, CondorcetFuse~\cite{condorcet_fuse}, and CoarsenRank~\cite{pan2022fast}. Besides, the listwise rank aggregation methods adopted are St.Agg~\cite{Stagg_alg} and CrowdAgg~\cite{crowd_agg_alg}, which are originally proposed for the partial rank aggregation scenarios. Therefore, to accommodate them to our setting, we apply these algorithms to each problem independently.
 Here, we provide a brief introduction of the adopted baseline methods:
 \begin{itemize}
    
    \item BradleyTerry~\cite{BradleyTerry}. It is a pairwise method that estimates the probability of the superior item within each pairwise comparison.
    
    \item CondorcetFuse~\cite{condorcet_fuse}. It is a pairwise method, which constructs the Condorcet Graph with $R$ items and derives the final rank via a Hamiltonian path. 

    \item CoarsenRank~\cite{pan2022fast}. It is specifically designed for mild model misspecification, and it assumes that the ideal preferences exist in a neighborhood of the actual preferences. Afterwards, it performs regular rank aggregations directly over a neighborhood of the preferences set.

    \item St.Agg~\cite{Stagg_alg}. It is a listwise method, which incorporates uncertainty into the aggregation process via introducing a prior distribution on ranks. Then it transforms the ranking functions to their expectations over this distribution.

    \item CrowdAgg~\cite{crowd_agg_alg}. It is an extension of the St.Agg method introduced above, where the annotator's quality information is further incorporated into the definition of the rank distribution.

\end{itemize}

For the comparing methods, hyperparameters are set as suggested in the original papers. For example, the hyperparameter $p$ in CrowdAgg is set to 0.95, as recommended by \cite{Stagg_alg}, and the rank measure adopted is $\kappa$-$\text{RBP}_s$. For St.Agg~\cite{Stagg_alg}, the ranking function is obtained by incorporating rank distribution into the mean position function. Moreover, for CoarsenRank~\cite{pan2022fast}, the optimal hyperparameter $\alpha$ is determined by the Deviance Information Criterion~\cite{DIC}.

\subsection{Data Synthetic Procedure}
In this section, we introduce the generation of the synthetic datasets in detail, including the construction of the confusion matrices and the generation process of the biased ranks provided by each annotator. 

To this end, we now take the $j$-th annotator as an example. First, for a given quality of annotators (namely, $e$) and a specific position $r\in \{1,2,\cdots, R\}$, we select a random scalar $v$ uniformly in the range of $[e, 1]$, which serves as the $(r,r)$-th element in $\Pi^{(j)}$. The value $v$ corresponds to the probability that the annotator $E_j$ gives the correct rank at the $r$-th position. Subsequently, the remaining $r-1$ positive values in the $r$-th row of $\Pi^{(j)}$ can be selected randomly, which must satisfy $\sum_{t=1,t\neq r}^R \Pi^{(j)}_{rt}=1-v$. Here, we use $\Pi^{(j)}_{rt}$ to denote the $(r,t)$-th element of $\Pi^{(j)}$. For clarity, here we take $R=3$ as an example. If $e=0.8$, a feasible quality matrix of an annotator is given by
\begin{equation}
\Pi^{(j)}=\begin{bmatrix}
     0.8 & 0.15 &  0.05\\
     0.05 & 0.9 &  0.05\\
     0.0 & 0.0 &  1.0\\
    \end{bmatrix}_{3 \times 3}.
\end{equation}
A reason for such a construction is that a higher annotator's quality $e$ is correlated with the annotator's superior ability. Therefore, this annotator is more likely to give correct ranks at each position. Reflected in the transition matrix, the diagonal values $\{\Pi_{tt}^{(j)}\}_{t=1}^R$ should be relatively large, which can be satisfied actually in our construction. Note that we introduce randomness for these diagonal values to ensure discrepancies between different annotators.

Subsequently, we elaborate on the generation process of the biased ranks based on $\Pi^{(j)}$. We denote the biased rank by $Rank_b$, which is blank in the beginning. The ground-truth rank (a random permutation) is denoted by $Rank_g$. First, we iterate through the array $[1,2,\cdots, R-1]$ sequentially. For a specific position $r$, we select a random value $p$ from the list $[1,2,\cdots, R]$ with probability $\Pi^{(j)}_{r\cdot}$. Such a process stops when we find a $p$ that satisfies $Rank_g[p]\not \in Rank_b$, and then we put $Rank_g[p]$ in the $r$-th position of $Rank_b$. Finally, the value in the $R$-th position is selected so that $Rank_b$ can be a permutation over $\{1,2,\cdots,R\}$. Therefore, we obtain one biased rank, and the other ranks are provided in the same manner.

\subsection{Experiments on Synthetic Datasets}

In this part, we compare the performance of our LAC with the baseline methods under various predefined conditions on the synthetic datasets introduced in the previous section. \par

There are mainly five factors associated with the synthetic datasets, \ie, the number of examples $I$, the length of rank $R$, the number of annotators $J$, the ability of annotators $e$, and the annotation ratio $\eta$. Here, the annotation ratio controls the ratio of annotators required to provide ranks for each problem. Notably, we set the basic parameters as $I=500$, $J=10$, $e=0.3$, $\eta=0.5$, and $R=5$ since our approach is applicable for some cases with a moderate number of problems and a suitable number of items to be ranked. Unless otherwise stated, this configuration is used throughout all our experiments.

\textbf{Performance on problems with different lengths.} We select the number of items in each problem, namely $R$, from the set $\{3, 4, 5, 6, 7\}$. Other parameters for the dataset are fixed as mentioned above, \ie, $I=500$, $J=10$, $e=0.3$, $\eta=0.5$. The accuracies, as well as the standard deviations over five independent trials of all methods, are shown in Table~\ref{result_on_different_R}, where the best record under each $R$ is in bold, and the second best record is underlined~(such presentation remains consistent throughout this paper). This table reveals that our LAC achieves relatively stable performance, demonstrating the superiority of our method on different lengths of problems over other baseline methods. It is worth noting that our LAC outperforms CrowdAgg by more than 4.7\% with various $R$, which suggests that explicitly modeling the annotators can obtain more satisfactory performance than tackling each problem independently.

\textbf{Performance on different numbers of examples.} We choose the number of examples $I$ within the set $\{100, 200, 300, 400, 500 \}$. The performance comparison of all methods is shown in Table~\ref{result_on_different_I}. Apparently, with the increase of $I$, our LAC still performs better than other baseline methods, which shows the excellence of LAC on various scales of the ranking problem. 

\textbf{Performance on different annotation ratios. } We select the annotation ratio $\eta$ in the range of $\{0.3, 0.4, 0.5, 0.6, 0.7\}$. The experimental results of all methods are shown in Table~\ref{result_on_different_eta}. As revealed in this table, when $\eta$ is very small, our LAC outperforms all the baseline methods by a large margin. Meanwhile, in all other cases, LAC shows superiority over the compared methods, especially when the annotation ratio is relatively large. 

\textbf{Performance on different numbers of annotators. } We pick the number of annotators $J$ from the set $\{10, 12, 14, 16, 18\}$ for each method. The detailed results are shown in Table~\ref{result_on_different_J}. As shown in the results, LAC consistently outperforms other methods in all tested scenarios, indicating its robust aggregation ability. Notably, with an increase in the number of annotators, our LAC generally obtains more accurate prediction. Moreover, our method is potentially applicable to some scenarios with a small number of annotators since our LAC surpasses the second best method CondorcetFuse by 13.90\% when $J=10$.

\textbf{Performance on different abilities of annotators. } We finally evaluate the performance of our LAC on different abilities of annotators, and thus we select $e$ in the range of $\{0.1, 0.3, 0.5, 0.7, 0.9\}$. The performance comparison is shown in Table~\ref{result_on_different_e}. This table reveals that, as the annotator's ability decreases to 0.1, LAC exhibits superior performance relative to other methods by a significant margin, which can be attributed to the characterization of the annotator's ability matrices. It further justifies that our method can handle low-quality annotations at different levels.

\setlength{\tabcolsep}{7.5pt}
\begin{table*}[t]
    \small
  \centering
  \caption{Performance comparison of various methods on synthetic dataset with different lengths of rank (in percent). The best record under each $R$ is in bold, and the second best record is underlined.}
  \begin{tabularx}{\textwidth}{lccccc}
    \toprule
  	Length of rank ($R$) &3&4&5&6&7\\ 
    \midrule
BradleyTerry~\cite{BradleyTerry}          & 80.91 $\pm$ 4.23 & \underline{79.50 $\pm$ 5.79}  & 74.17 $\pm$ 3.81 & 76.48 $\pm$ 2.23   &  77.54 $\pm$ 4.69  \\
CondorcetFuse~\cite{condorcet_fuse}      & 73.94 $\pm$ 4.28 & 75.65 $\pm$ 4.40 & \underline{75.62 $\pm$ 4.87} & \underline{79.70 $\pm$ 1.90}  & \underline{82.89 $\pm$ 2.15}   \\
CoarsenRank~\cite{pan2022fast}             & 67.05 $\pm$ 7.96 & 69.49 $\pm$ 7.30 & 67.03 $\pm$ 3.92 & 71.83 $\pm$ 3.34  &  73.80 $\pm$ 4.60  \\
St.Agg~\cite{Stagg_alg}             & 77.95 $\pm$ 7.53 & 60.65 $\pm$ 3.58 & 71.57 $\pm$ 2.46 & 59.40 $\pm$ 0.52  &  77.08 $\pm$ 3.59  \\
CrowdAgg~\cite{crowd_agg_alg}             & \underline{81.38 $\pm$ 5.90} & 63.53 $\pm$ 3.20 & 72.64 $\pm$ 2.26 & 64.47 $\pm$ 0.59  &  78.49 $\pm$ 3.44  \\
LAC               & \textbf{86.08 $\pm$ 3.84} & \textbf{88.09 $\pm$ 4.53} & \textbf{89.20 $\pm$ 3.73} &  \textbf{92.84 $\pm$ 1.57} & \textbf{94.72 $\pm$ 2.43}   \\ 
    \bottomrule
  \end{tabularx}

  \label{result_on_different_R}

\end{table*}

\setlength{\tabcolsep}{6.0pt}
\begin{table*}[t]
  \centering
  \caption{Performance comparison of various methods on synthetic dataset with different example numbers (in percent). The best record under each $I$ is in bold, and the second best record is underlined.}
  \small
  \begin{tabularx}{\textwidth}{lccccc}
    \toprule
       Number of problems ($I$) &100&200&300&400&500\\ 
    \midrule
BradleyTerry~\cite{BradleyTerry}  & \underline{81.08 $\pm$ 4.27} & \underline{79.74 $\pm$ 3.52} & \underline{81.25 $\pm$ 3.75} & \underline{80.25 $\pm$ 3.42}  & \underline{81.53 $\pm$ 2.48}    \\         
CondorcetFuse~\cite{condorcet_fuse}      & 75.72 $\pm$ 3.33  & 77.38 $\pm$ 3.11  & 76.10 $\pm$ 3.27 & 77.06 $\pm$ 2.77   & 75.88 $\pm$ 3.27    \\
CoarsenRank~\cite{pan2022fast}             & 64.76 $\pm$ 3.77 & 66.24 $\pm$ 5.31 & 66.08 $\pm$ 5.06 & 67.13 $\pm$ 4.68   &  67.03 $\pm$ 3.92  \\
St.Agg~\cite{Stagg_alg}             & 68.96 $\pm$ 4.00 & 71.32 $\pm$ 4.66 & 70.93 $\pm$ 4.99 & 71.50 $\pm$ 4.75  &  71.20 $\pm$ 4.00  \\
CrowdAgg~\cite{crowd_agg_alg}             & 70.56 $\pm$ 3.81 & 72.58 $\pm$ 3.94 & 72.36 $\pm$ 4.28 & 73.15 $\pm$ 4.35  &  72.72 $\pm$ 3.14  \\
LAC                 & \textbf{95.96 $\pm$ 1.71} & \textbf{96.04 $\pm$ 2.08} & \textbf{96.14 $\pm$ 1.92} & \textbf{96.03 $\pm$ 2.32}  &  \textbf{96.33 $\pm$ 1.91}  \\ 
    \bottomrule
  \end{tabularx}
  \label{result_on_different_I}

\end{table*}

\setlength{\tabcolsep}{7.0pt}
\begin{table*}[t]
  \centering
  \caption{Performance comparison of various methods on synthetic dataset with different annotation ratios (in percent). The best record under each $\eta$ is in bold, and the second best record is underlined.}
  \small
  \begin{tabularx}{\textwidth}{lccccc}
    \toprule
  	Annotation ratio ($\eta$) &0.3&0.4&0.5&0.6&0.7\\ 
    \midrule
BradleyTerry~\cite{BradleyTerry}                & 74.17 $\pm$ 3.81  & \underline{78.47 $\pm$ 2.97}  & \underline{81.53 $\pm$ 2.48}  & 83.89 $\pm$ 2.56  & 85.09 $\pm$ 3.19    \\ 
CondorcetFuse~\cite{condorcet_fuse}            & \underline{75.62 $\pm$ 4.87}  & 76.56 $\pm$ 2.28  & 75.88 $\pm$ 3.27  & 76.48 $\pm$ 3.06  & 76.00 $\pm$ 3.78    \\ 
CoarsenRank~\cite{pan2022fast}                   & 67.03 $\pm$ 3.92  & 70.23 $\pm$ 3.71  & 72.57 $\pm$ 2.42  & 76.62 $\pm$ 2.79  & 78.31 $\pm$ 3.22   \\ 
St.Agg~\cite{Stagg_alg}             & 71.20 $\pm$ 4.00 & 72.88 $\pm$ 2.94 & 78.88 $\pm$ 3.03 & 82.26 $\pm$ 2.80  &  84.98 $\pm$ 3.27  \\
CrowdAgg~\cite{crowd_agg_alg}             & 72.72 $\pm$ 3.14 & 77.47 $\pm$ 3.75 & 81.36 $\pm$ 3.62 & \underline{86.17 $\pm$ 2.18}  & \underline{87.02 $\pm$ 3.41} \\
LAC                   & \textbf{89.20 $\pm$ 3.73}  & \textbf{93.00 $\pm$ 2.09}   & \textbf{96.33 $\pm$ 1.91}  & \textbf{97.88 $\pm$ 1.07}  &  \textbf{98.60 $\pm$ 0.91}   \\ 
    \bottomrule
  \end{tabularx}
  \label{result_on_different_eta}
\end{table*}

\setlength{\tabcolsep}{5.3pt}
\begin{table*}[t]

  \centering
  \caption{Performance comparison of various methods on synthetic dataset with different numbers of annotators (in percent). The best record under each $J$ is in bold, and the second best record is underlined.}
  \small
  \begin{tabularx}{\textwidth}{lccccc}
    \toprule
  	Number of annotators ($J$) &10&12&14&16&18\\ 
    \midrule
BradleyTerry~\cite{BradleyTerry}                &  74.06 $\pm$ 1.93 & 72.22 $\pm$ 1.94  & \underline{79.29 $\pm$ 0.57}  & \underline{79.34 $\pm$ 1.43}  &  81.72 $\pm$ 0.62   \\ 
CondorcetFuse~\cite{condorcet_fuse}            &  \underline{76.85 $\pm$ 2.42} & \underline{76.64 $\pm$ 1.23}  & 78.30 $\pm$ 0.59  & 76.34 $\pm$ 0.45  &  75.42 $\pm$ 1.68   \\ 
CoarsenRank~\cite{pan2022fast}                  &  68.05 $\pm$ 1.68 & 66.41 $\pm$ 1.61  & 71.41 $\pm$ 0.38  & 70.96 $\pm$ 1.66  &  73.38 $\pm$ 1.04  \\ 
St.Agg~\cite{Stagg_alg}             & 71.57 $\pm$ 2.46 & 70.25 $\pm$ 1.98 & 72.89 $\pm$ 1.52 & 73.37 $\pm$ 2.41  &  79.40 $\pm$ 2.08  \\
CrowdAgg~\cite{crowd_agg_alg}             & 72.64 $\pm$ 2.26 & 71.35 $\pm$ 2.30 & 77.75 $\pm$ 1.73 & 77.39 $\pm$ 2.40  &  \underline{82.52 $\pm$ 1.73}  \\
LAC            &  \textbf{90.75 $\pm$ 1.01} & \textbf{90.50 $\pm$ 0.74}  & \textbf{95.12 $\pm$ 0.61}  &  \textbf{94.37 $\pm$ 0.80} & \textbf{96.85 $\pm$ 1.15}    \\ 
    \bottomrule
  \end{tabularx}
  \label{result_on_different_J}
\end{table*}

\begin{table*}[!t]
  \centering
  \caption{Performance comparison of various methods on synthetic dataset with different abilities of annotators (in percent). The best record under each $e$ is in bold, and the second best record is underlined.}
  \small
  \begin{tabularx}{\textwidth}{lccccc}
    \toprule
  	Ability of annotators ($e$) &0.1&0.3&0.5&0.7&0.9\\ 
    \midrule
BradleyTerry~\cite{BradleyTerry}                & 53.82 $\pm$ 5.52  & 64.29 $\pm$ 4.84  & 74.17 $\pm$ 3.81  & 84.74 $\pm$ 2.79  &  94.82 $\pm$ 0.62    \\ 
CondorcetFuse~\cite{condorcet_fuse}           & \underline{56.90 $\pm$ 6.22}  & \underline{65.77 $\pm$ 4.90}  & \underline{75.62 $\pm$ 4.87}  & \underline{86.82 $\pm$ 3.39} &  \underline{95.35 $\pm$ 0.71}   \\ 
CoarsenRank~\cite{pan2022fast}               & 47.55 $\pm$ 5.16  &  56.81 $\pm$ 3.95 & 67.03 $\pm$ 3.92  & 79.12 $\pm$ 4.57  & 91.64 $\pm$ 1.44     \\ 
St.Agg~\cite{Stagg_alg}             & 51.12 $\pm$ 5.44 & 61.22 $\pm$ 4.06 & 71.20 $\pm$ 4.00 & 82.68 $\pm$ 3.82  &  93.50 $\pm$ 1.55  \\
CrowdAgg~\cite{crowd_agg_alg}             & 54.32 $\pm$ 5.82 & 63.41 $\pm$ 3.71 & 72.72 $\pm$ 3.14 & 83.26 $\pm$ 3.67  &  93.79 $\pm$ 1.46  \\
LAC            & \textbf{68.41 $\pm$ 7.54}  & \textbf{79.85 $\pm$ 5.28}  & \textbf{89.20 $\pm$ 3.73}  &  \textbf{96.32 $\pm$ 1.91} & \textbf{99.60 $\pm$ 0.19}    \\ 	
    \bottomrule
  \end{tabularx}
  \label{result_on_different_e}
\end{table*}

\begin{figure}[t]
  \centering
    \centering
      {\includegraphics[width=0.5\linewidth]{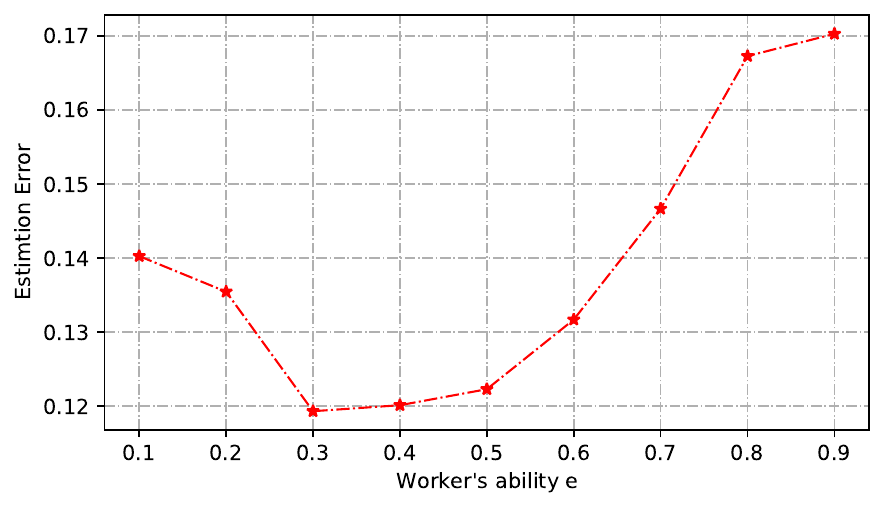}}
    \caption{Overall estimation error of LAC with various $e$.} 
	\label{overall_estimation_error}
\end{figure}

\begin{figure*}[th]
    \centering
	\begin{minipage}{1\linewidth}
		\centering
		\subfigure[$e=0.1$]
		{\includegraphics[width=0.31\linewidth]{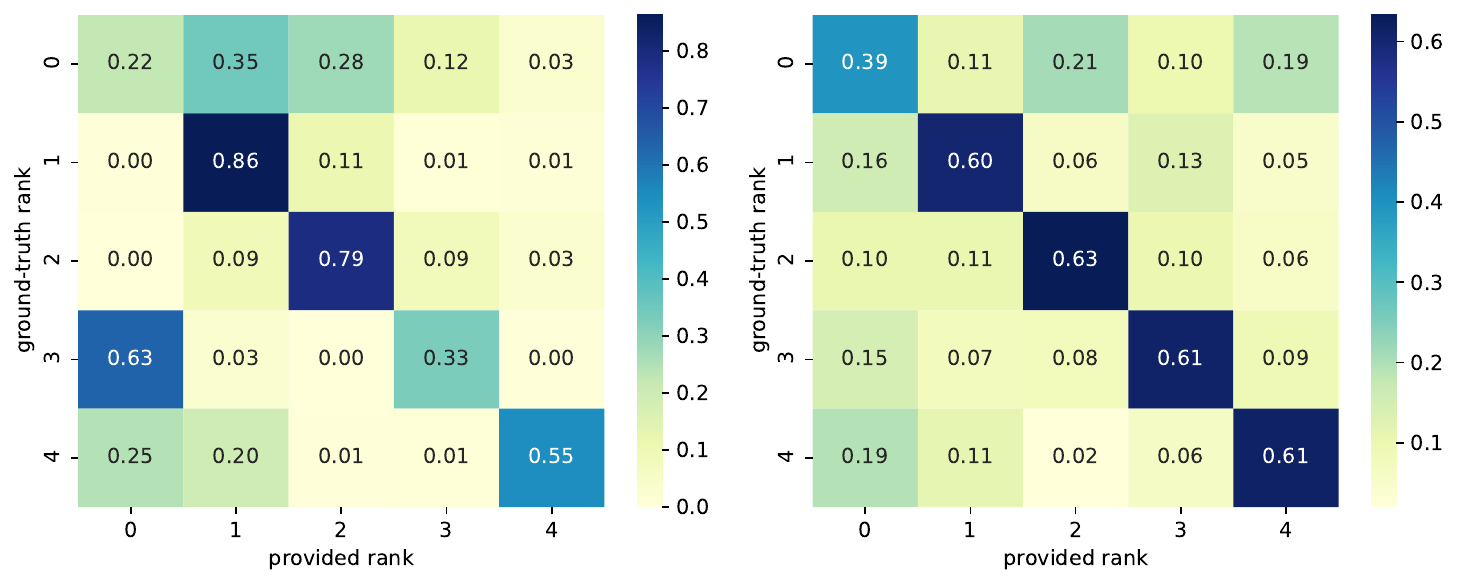}}
		\subfigure[$e=0.2$]
		{\includegraphics[width=0.31\linewidth]{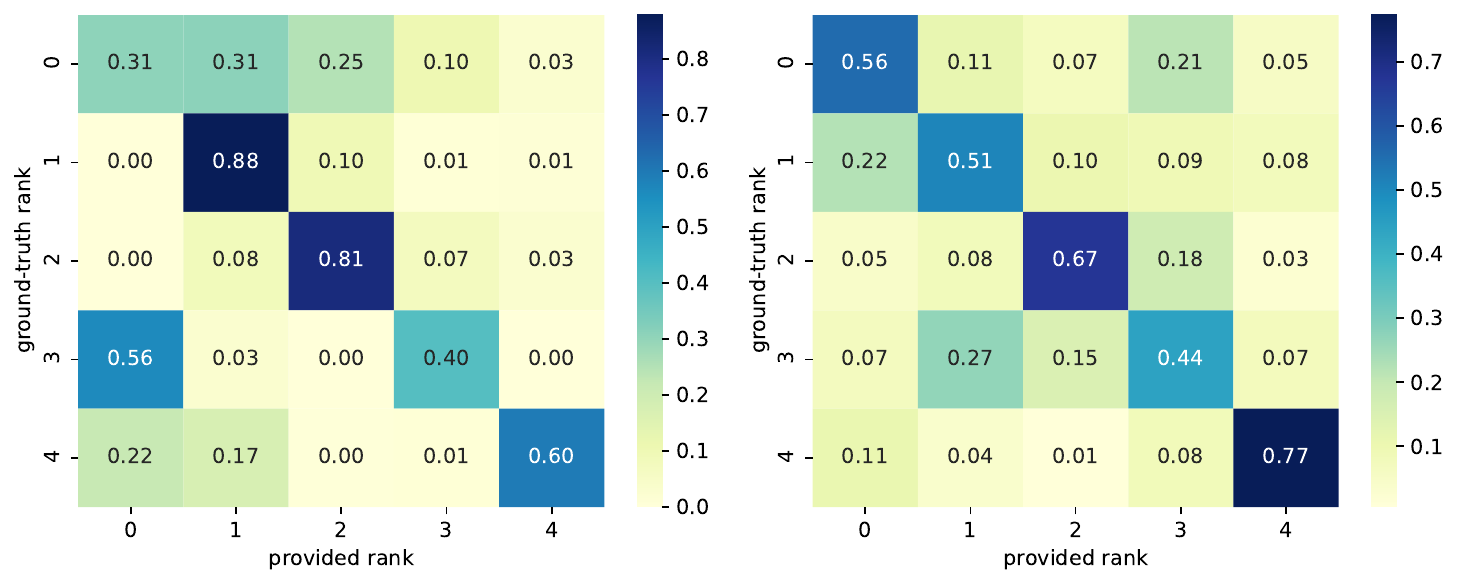}}
            \subfigure[$e=0.3$]
		{\includegraphics[width=0.31\linewidth]{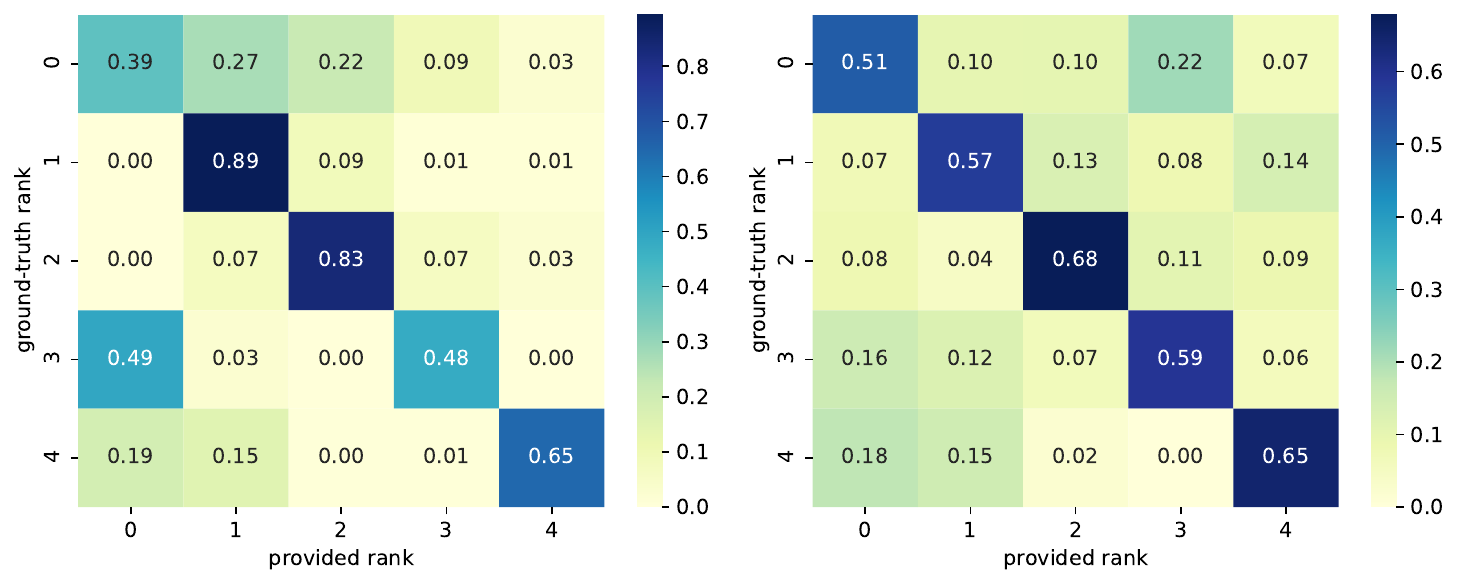}}
	\end{minipage}
        \begin{minipage}{1\linewidth}
		\centering
		\subfigure[$e=0.4$]
		{\includegraphics[width=0.31\linewidth]{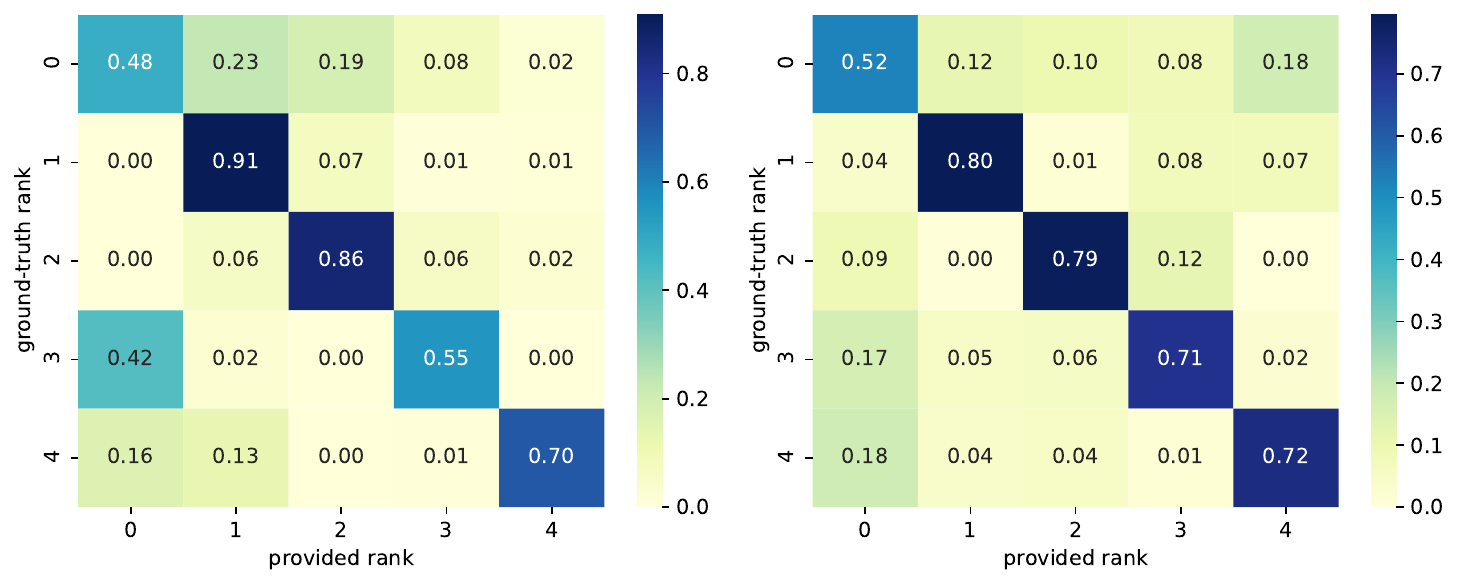}}
            \subfigure[$e=0.5$]
		{\includegraphics[width=0.31\linewidth]{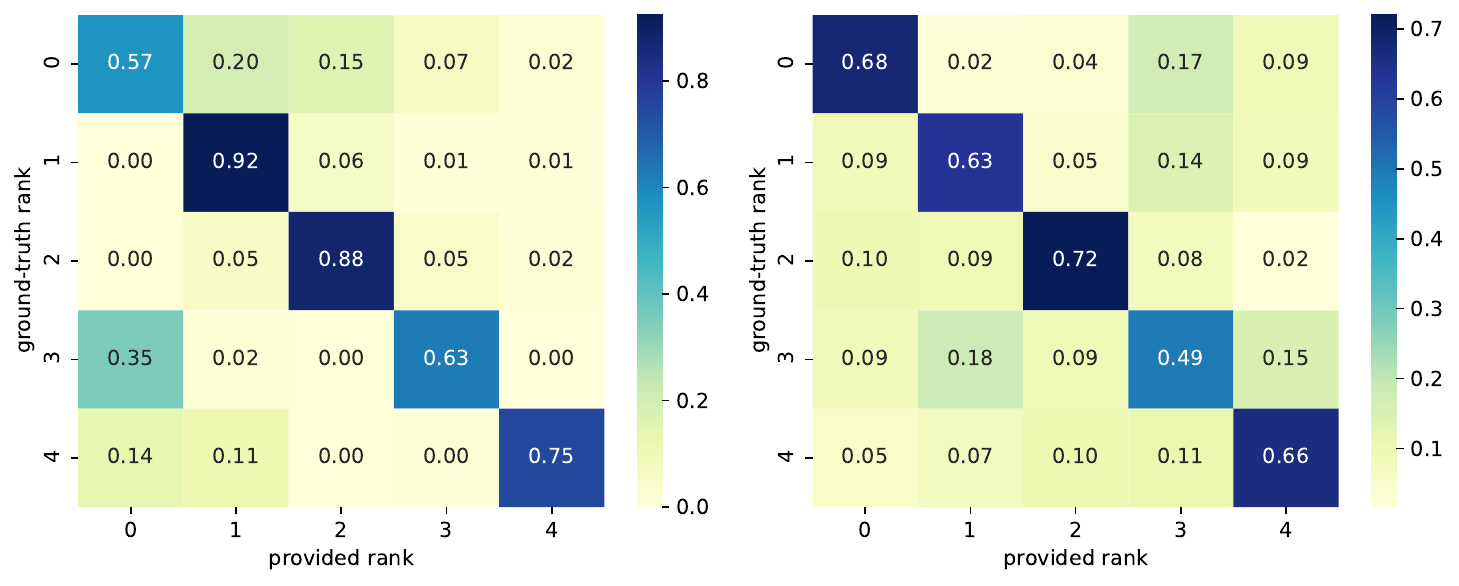}}
		\subfigure[$e=0.6$]
		{\includegraphics[width=0.31\linewidth]{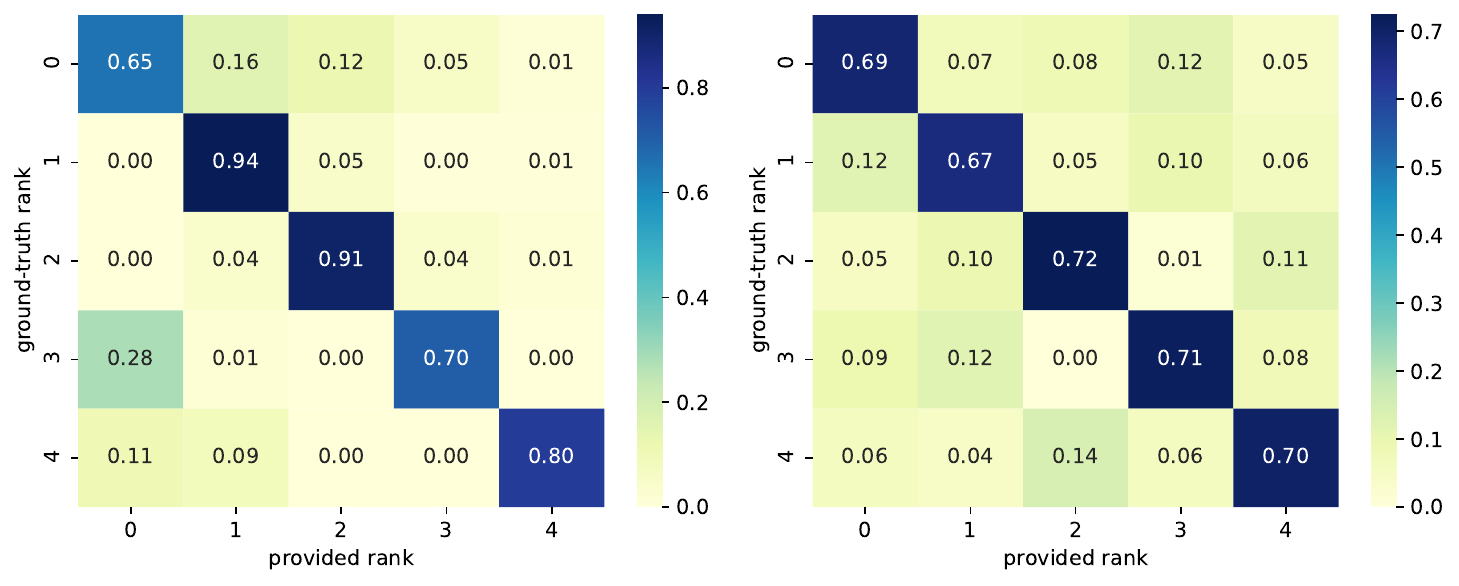}}
	\end{minipage}
        
        \begin{minipage}{1\linewidth}
		\centering
		\subfigure[$e=0.7$]
		{\includegraphics[width=0.31\linewidth]{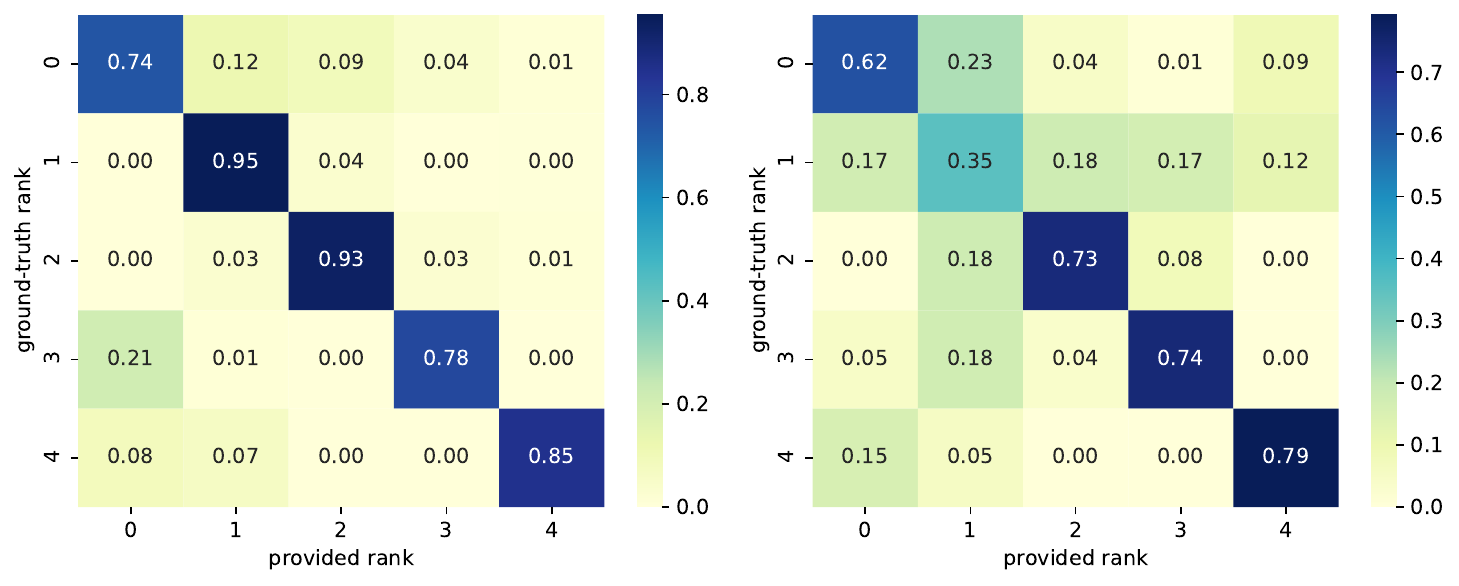}}
		\subfigure[$e=0.8$]
		{\includegraphics[width=0.31\linewidth]{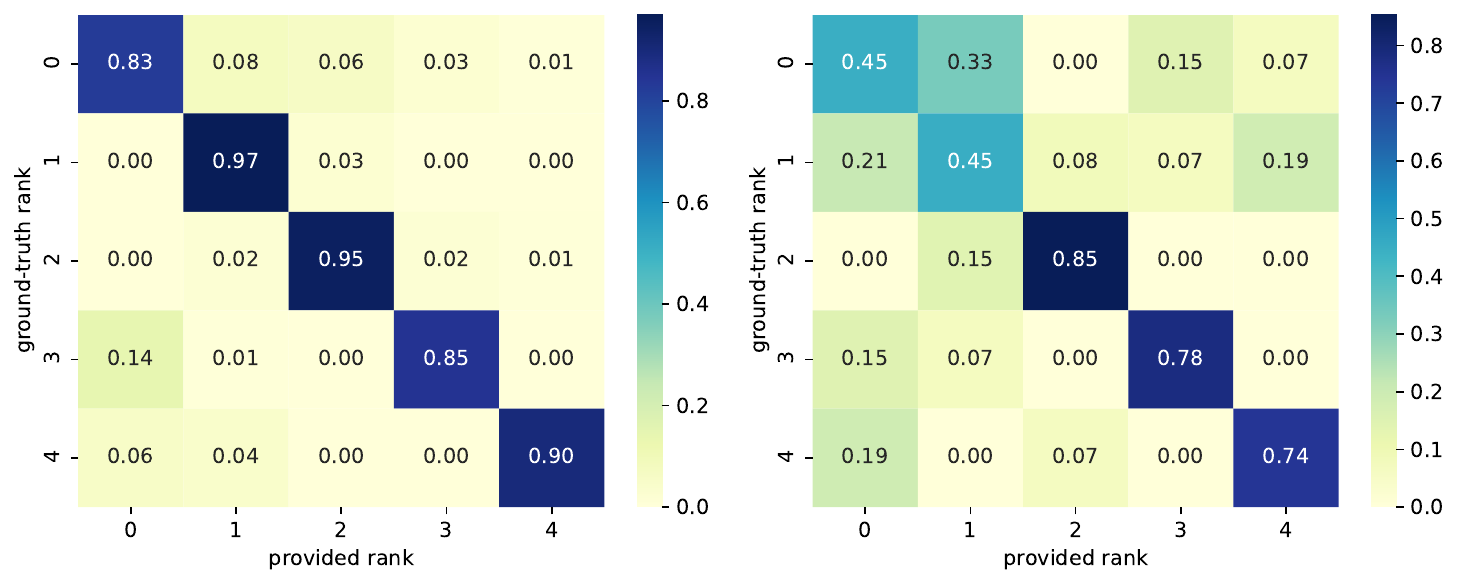}}
            \subfigure[$e=0.9$]
		{\includegraphics[width=0.31\linewidth]{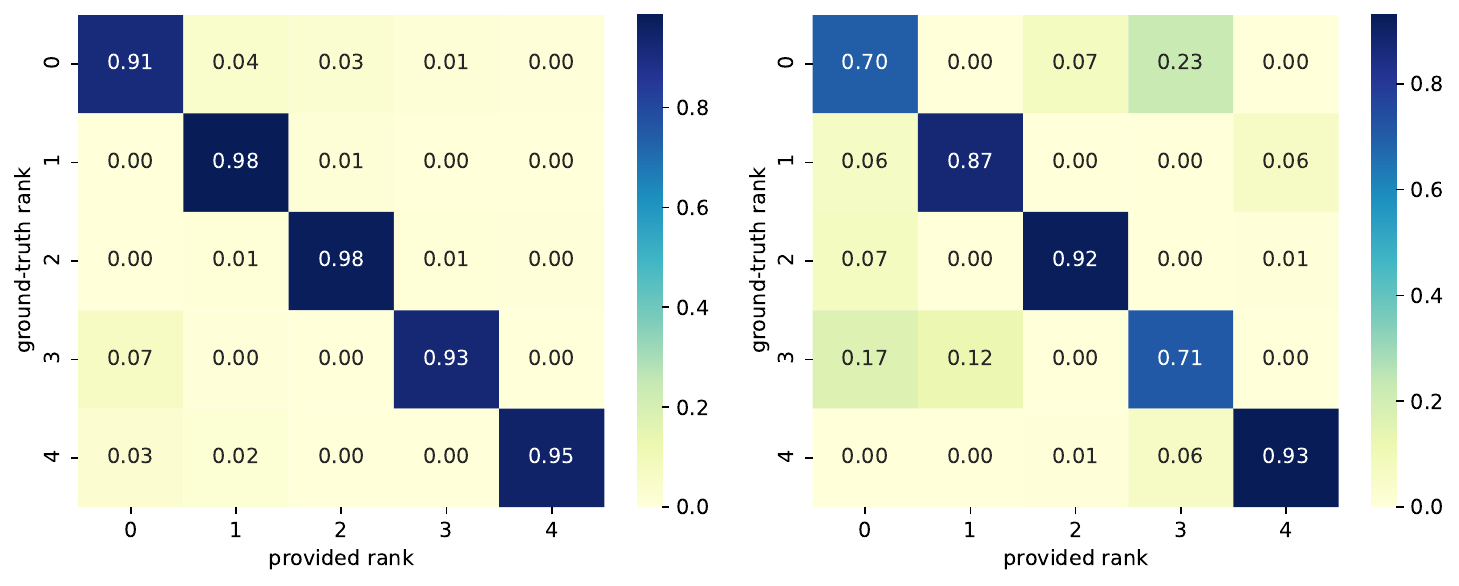}}
	\end{minipage}
        \caption{Comparisons between the ground-truth ability matrices of annotators (left) and the estimated ones (right), which illustrate that our estimated confusion matrices are similar to the corresponding ground-truth ones in most cases.} 
	\label{comparison}
\end{figure*}

\begin{figure*}[t]
	\begin{minipage}{1\linewidth}
		\centering
		\subfigure[Different numbers of examples]
		{\includegraphics[width=0.48\linewidth]{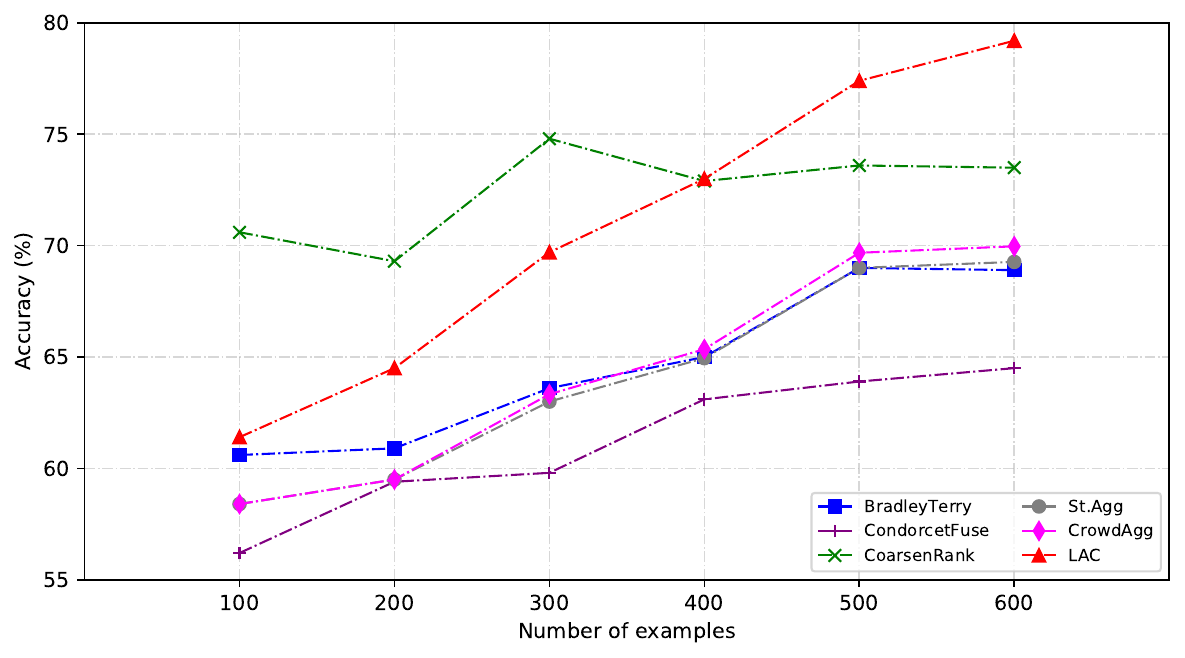}}
            \hfill
		\subfigure[Different numbers of annotators]
		{\includegraphics[width=0.48\linewidth]{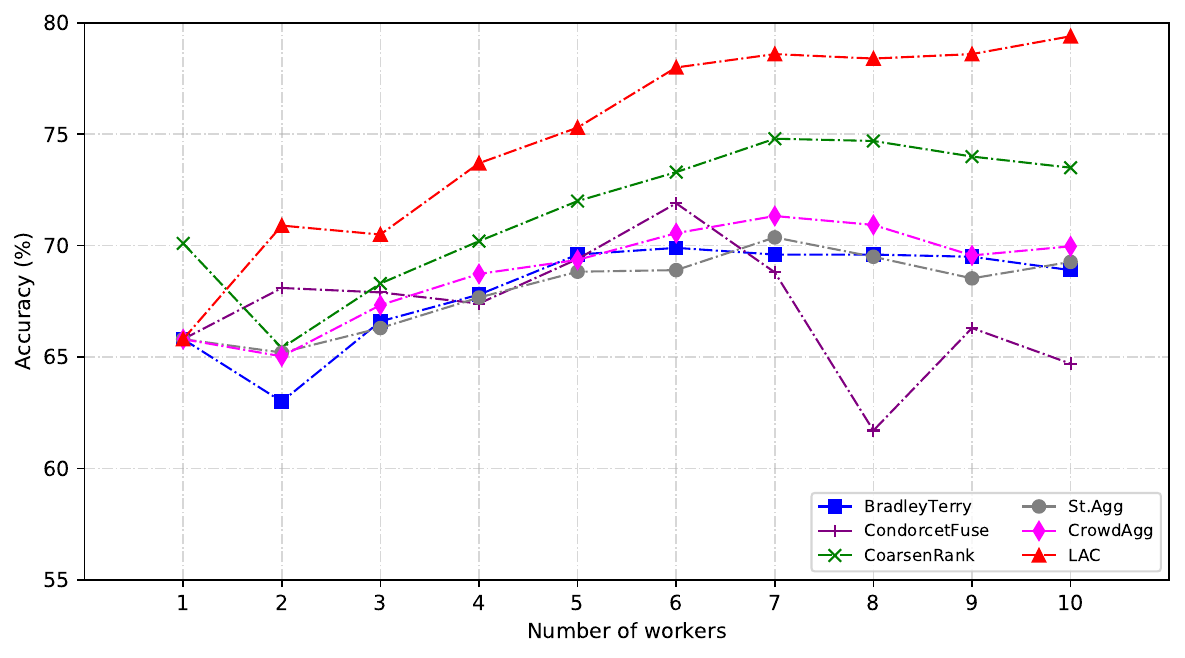}}
	\end{minipage} 
    \caption{Test accuracy curves on ParaRank with different numbers of examples and different numbers of annotators for all the compared methods.} 
	\label{ParaRank_dataset}
\end{figure*}

\textbf{Estimation error on the ability matrices of annotators.} Since LAC explicitly models the annotators' quality via confusion matrices, we also carry out experiments to verify the effectiveness of our method in estimating such matrices. To this end, we analyze the estimation errors quantitatively, and we propose to use the following metric to calculate the overall estimation error for a given $e$:
\begin{equation}
    {Error}=\frac{1}{R^2\cdot J}\sum_{j=1}^J\sum_{t,k=1}^R \left|\Pi_{tk}^{(j)}-\Pi_{tk}^{(j),est}\right|,
\end{equation}
where $\Pi^{(j),est}$ is the estimated ability matrix, and $\Pi^{(j)}$ is the ground-truth matrix for the annotator $E_j$. The estimation errors with various $e$ are shown in Fig.~\ref{overall_estimation_error}. From this figure, we identify that our estimations achieve small errors in most cases, demonstrating the effectiveness of EM steps in finding the optimal values of latent variables. Note that when $e>0.3$, the estimation error exhibits an increasing trend as $e$ increases. This can be attributed to the fact that a larger $e$ leads to a sparser ability matrix (when $e=1$, $\Pi^{(j)}$ degenerates to an identity matrix, representing the sparsest case), and thus it is harder to estimate such a matrix accurately.

Additionally, more qualitative results are provided. Here, we showcase some comparisons between the ground-truth matrices and the estimated ones, which are illustrated in Fig.~\ref{comparison}. In this figure, for simplicity, we only present the results related to the first annotator. Obviously, our estimations are very close to the ground-truth matrices in most cases, which implies that our estimation is reliable.

To summarize, the experimental results on synthetic datasets clearly indicate that our LAC can obtain more accurate estimations of ground-truth ranks than other methods. Besides, our LAC performs well in estimating the annotator's ability matrices. Next, we conduct experiments on a real-world dataset to further verify its superior performance.

\begin{table}[!t]
\small
    \centering
  \caption{Basic properties of the collected ParaRank dataset.}
  \begin{tabular}{lc}
    \toprule
  	 \multicolumn{1}{c}{Property} & \multicolumn{1}{c}{Number}\\
    \midrule
    \rule{0pt}{1pt}
	Paragraphs\,/\,Problems	&   \multicolumn{1}{c}{600} \\
    \rule{0pt}{8.5pt}
        Words per paragraph & \multicolumn{1}{c}{300} \\
    \rule{0pt}{8.5pt}
        Sentences per paragraph & \multicolumn{1}{c}{5} \\
    \rule{0pt}{8.5pt}
        Crowdsourced annotators & \multicolumn{1}{c}{25} \\
    \rule{0pt}{8.5pt}
        Problems for each annotator & \multicolumn{1}{c}{239} \\
    \rule{0pt}{8.5pt}
        Total annotations & \multicolumn{1}{c}{5,981} \\
    \bottomrule
  \end{tabular}
  \label{tab: properties_of_pararank}
\end{table}

\subsection{Experiments on Real-world Dataset}

There are requirements for listwise full rank aggregation in real-world business. Specifically, in the game business of NetEase, the large language models~\cite{LLM} (LLMs) can automatically generate plots and quests for players. However, the logic of the content generated by LLMs sometimes cannot align with the human logic. Therefore, we have to reorder the sentences of the generated paragraph to make it more reasonable. Subsequently, the reordered sentences and the originally generated sentences are used to train a reward model, which is further employed to fine-tune the LLM. However, hiring experts to determine the order of sentences is costly and inefficient, and thus we post this task on the crowdsourcing platforms to collect redundant annotations. Here, we selected some examples of paragraphs and collected a real-world dataset called ``ParaRank"\footnote{The ParaRank dataset is publicly available at \url{https://anonymous.4open.science/r/LAC-B871}.}, to evaluate the proposed rank aggregation method. 

The basic properties of ParaRank are listed in Table.~\ref{tab: properties_of_pararank}. As shown in this table, the ParaRank dataset comprises 600 paragraphs generated by LLMs, with 300 words per paragraph on average, and each paragraph contains five sentences. Here, each paragraph corresponds to a ranking problem, where the sentences within it need to be ranked. We posted all the paragraphs on the NetEase Youling crowdsourcing platform\footnote{URL: \url{https://zhongbao-web-9109-80.apps-fp.danlu.netease.com/mark/task}} to obtain noisy rankings for them. In more detail, there are a total of 25 crowdsourced annotators, each with a satisfactory track record of historical accuracy.  We solicited their recommendations for the most suitable sequence of the five sentences in each paragraph. Each problem was annotated by ten different annotators, and on average, each annotator annotated 239 problems. In terms of expenses, each annotator was paid 2 RMB for each problem, resulting in a total cost of 20,000 RMB. In terms of time cost, the average time expended for a single annotation was about 24.2 second, and the entire task was completed in 4 days. Finally, we obtained a total of 5,981 annotations for 600 problems. For evaluation, we collected the ground-truth rank for each problem, which was provided manually by human experts.

Five adopted baseline methods and our LAC are evaluated on ParaRank. For parameter setting, we choose the parameters when a method reaches its best performance on the validation set. For example, the hyperparameter $p$ in CrowdAgg is set to 0.95, and the rank measure adopted is $\kappa$-$\text{RBP}_s$. 
The results are presented in Table~\ref{result_on_real_world_dataset}. Based on the experimental results, our LAC has demonstrated superior performance to all other methods. To further investigate the performance of all methods on this real-world dataset, we gradually increase the number of examples (or the number of annotators) and get the corresponding test accuracy. The accuracy curves are illustrated in Fig.~\ref{ParaRank_dataset}. The two graphs show that most methods achieve better performance with the increase in number of examples and annotators, but LAC has more significant performance gains than the other methods. This is because LAC models the difficulty of each problem and the ability of the annotators in a more detailed manner, which results in enhanced robustness compared to other methods that do not explicitly address these critical factors.

\setlength{\tabcolsep}{7.9pt}
\begin{table*}[!t]
  \centering
  \footnotesize
  \caption{Performance comparison of Five baseline methods and our LAC on real-world dataset (in percent).}
  \begin{tabularx}{\textwidth}{cccccc}
    \toprule
  	BradleyTerry~\cite{BradleyTerry} & CondorcetFuse~\cite{condorcet_fuse} & CoarsenRank~\cite{pan2022fast} & St.Agg~\cite{Stagg_alg} & CrowdAgg~\cite{crowd_agg_alg} & LAC\\ 
    \midrule
	  	68.93   &  64.53 & \underline{73.52}  & 69.27 & 69.97 & \textbf{79.26}  \\
    \bottomrule
  \end{tabularx}
  \label{result_on_real_world_dataset}
\end{table*}

\section{Conclusion}

\noindent In this paper, we propose a novel rank aggregation method dubbed LAC to deal with the listwise input in crowdsourcing. Unlike previous listwise methods that may only consider the partial ranks across items, our LAC delves into the underexplored problem of full rank aggregation. Moreover, LAC incorporates both the ability of annotators and the difficulty of problems into the modeling by introducing two sets of confusion matrices. Such matrices and the true ranks can be deduced iteratively by the EM algorithm. To our knowledge, LAC is the first work to directly deal with the full rank aggregation problem in listwise crowdsourcing, and simultaneously infer the difficulty of problems, the ability of annotators, and the ground-truth ranks in an unsupervised way. To evaluate our method on the listwise full rank aggregation task, we collect a dataset with real-world business consideration. Experimental results on both synthetic and real-world datasets demonstrate the effectiveness of our proposed LAC method.

\begin{acks}
\noindent This work is supported by the National Natural Science Foundation Program of China under Grants U1909207, U21B2029, and 62336003, the Natural Science Foundation of Jiangsu Province under Grant BK20220080, and the Key R\&D Program of Zhejiang Province under Grant No. 2022C01011.
\end{acks}

\bibliographystyle{ACM-Reference-Format}
\bibliography{references}


\end{document}